\documentclass[10pt,twocolumn,letterpaper]{article}

\usepackage{iccv}              

\usepackage[accsupp]{axessibility}

\usepackage{algorithm}
\usepackage{algpseudocode}
\usepackage{setspace}
\usepackage{multirow}
\usepackage{amsthm}
\usepackage{graphicx}
\usepackage{listings}
\usepackage{tabularx}

\usepackage[table]{xcolor}
\definecolor{LightCyan}{gray}{0.9}

\newcommand{\red}[1]{{\color{red}#1}}
\newcommand{\blue}[1]{{\color{blue}#1}}

\usepackage{amsmath}

\usepackage{algorithm}
\usepackage{algpseudocode}

\usepackage[utf8]{inputenc} 
\usepackage[T1]{fontenc}    
\usepackage{url}            
\usepackage{booktabs}       
\usepackage{amsfonts}       
\usepackage{nicefrac}       
\usepackage{microtype}      
\usepackage{xcolor}       
\usepackage{xkcdcolors}
\usepackage{times}
\usepackage{epsfig}
\usepackage{graphicx}
\usepackage{placeins}
\usepackage{multirow}
\usepackage{caption}
\usepackage[skip=5pt]{caption}
\usepackage{rotating}
\usepackage{cancel}
\usepackage{setspace}

\usepackage{bm}
\usepackage{bibunits} 

\usepackage{amssymb,amsthm}
\usepackage{amsmath}
\usepackage{graphicx}
\usepackage{wrapfig,lipsum,booktabs}

\usepackage{epsfig}
\usepackage{tikz}
\usetikzlibrary{spy}
\usepackage{algpseudocode}
\usepackage{algorithm}
\usepackage{mathrsfs}

\usepackage{nicefrac}       
\usepackage{booktabs}       

\usepackage{thmtools,thm-restate}
\usepackage{listings}
\usepackage{lstautogobble}  %
\usepackage{color}          %
\usepackage{zi4}

\usepackage{indentfirst}

\usepackage{makecell}
\usepackage{tabularx}
\usepackage{animate}
\usepackage{capt-of}

\usepackage{colortbl} 
\usepackage[table]{xcolor}
\definecolor{mylightgray}{RGB}{236, 236, 236}

\usepackage{amsmath,amsfonts,bm}

\def\eqref#1{Eq.~\ref{#1}}

\def\1{\bm{1}}

\def\rmI{{\mathbf{I}}}

\def\va{{\bm{a}}}

\def\vc{{\bm{c}}}

\def\vp{{\bm{p}}}

\def\vr{{\bm{r}}}
\def\vs{{\bm{s}}}

\def\vu{{\bm{u}}}
\def\vv{{\bm{v}}}

\def\vx{{\bm{x}}}

\def\mI{{\bm{I}}}

\DeclareMathAlphabet{\mathsfit}{\encodingdefault}{\sfdefault}{m}{sl}
\SetMathAlphabet{\mathsfit}{bold}{\encodingdefault}{\sfdefault}{bx}{n}

\newcommand{\f}{{\boldsymbol f}}

\newcommand{\x}{{\boldsymbol x}}

\newcommand{\z}{{\boldsymbol z}}

\newcommand{\epsilonb}{{\boldsymbol \epsilon}}
\newcommand{\mub}{{\boldsymbol \mu}}

\newcommand{\Ed}{{\mathbb E}}

\newcommand{\Nc}{{\mathcal N}}

\usepackage{capt-of}
\usepackage{diagbox}

\definecolor{C0}{rgb}{0.121569, 0.466667, 0.705882}
\definecolor{C1}{rgb}{1.000000, 0.498039, 0.054902}
\definecolor{C2}{rgb}{0.172549, 0.627451, 0.172549}
\definecolor{C3}{rgb}{0.839216, 0.152941, 0.156863}
\definecolor{C4}{rgb}{0.580392, 0.403922, 0.741176}
\definecolor{C5}{rgb}{0.549020, 0.337255, 0.294118}
\definecolor{C6}{rgb}{0.890196, 0.466667, 0.760784}
\definecolor{C7}{rgb}{0.498039, 0.498039, 0.498039}
\definecolor{C8}{rgb}{0.737255, 0.741176, 0.133333}
\definecolor{C9}{rgb}{0.090196, 0.745098, 0.811765}
\definecolor{trolleygrey}{rgb}{0.5, 0.5, 0.5}

\definecolor{BrickRed}{rgb}{0.6,0,0}
\definecolor{RoyalBlue}{rgb}{0,0,0.8}
\definecolor{Tdgreen}{rgb}{0,0.4,0.7}
\definecolor{pinegreen}{rgb}{0.0, 0.47, 0.44}
\definecolor{cornellred}{rgb}{0.7, 0.11, 0.11}
\definecolor{cadmiumgreen}{rgb}{0.0, 0.42, 0.24}
\definecolor{spirodiscoball}{rgb}{0.06, 0.75, 0.99}
\definecolor{mylightblue}{rgb}{0.85, 0.90, 0.94}
\definecolor{maroon}{cmyk}{0,0.87,0.68,0.32}

\usepackage{pifont}
\definecolor{iccvblue}{rgb}{0.21,0.49,0.74}
\usepackage[pagebackref,breaklinks,colorlinks,allcolors=iccvblue]{hyperref}

\def\paperID{11787} 
\def\confName{ICCV}
\def\confYear{2025}
\def\eqref#1{Eq.~(\ref{#1})}

\title{Free$^2$Guide: Training-Free Text-to-Video Alignment using Image LVLM 
}

\author{Jaemin Kim,$\quad \quad \quad$Bryan Sangwoo Kim,$\quad \quad \quad$Jong Chul Ye\\
Kim Jaechul Graduate School of AI, KAIST\\
{\tt\small \{kjm981995,  bryanswkim,  jong.ye\}@kaist.ac.kr}
}

\begin{document}


\twocolumn[{%
\renewcommand\twocolumn[1][]{#1}%
\maketitle
\vspace{-0.7cm}
\begin{center}
    \newcommand{\numColumns}{4}
    \newcommand{\columnSpacing}{0.1cm}
    
    \begin{tabularx}{\textwidth}{XXXX}
        \centering \small \textbf{Baseline} & 
        \centering \small \textbf{Free$^2$Guide} & 
        \centering \small \textbf{Baseline} &
        \centering \small \textbf{Free$^2$Guide}
    \end{tabularx}

    \begin{tabular}{
        @{}
        p{\dimexpr(\textwidth-\columnSpacing*(\numColumns-1))/\numColumns} @{\hspace{\columnSpacing}}
        p{\dimexpr(\textwidth-\columnSpacing*(\numColumns-1))/\numColumns} @{\hspace{\columnSpacing}}
        p{\dimexpr(\textwidth-\columnSpacing*(\numColumns-1))/\numColumns} @{\hspace{\columnSpacing}}
        p{\dimexpr(\textwidth-\columnSpacing*(\numColumns-1))/\numColumns} @{}
    }
        \includegraphics[width=\linewidth]{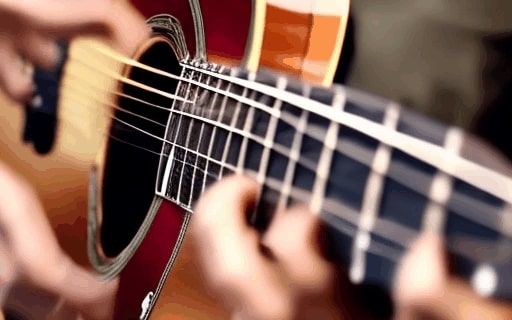} &
        \includegraphics[width=\linewidth]{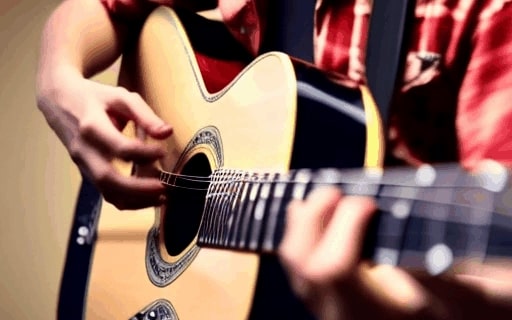} &
        \includegraphics[width=\linewidth]{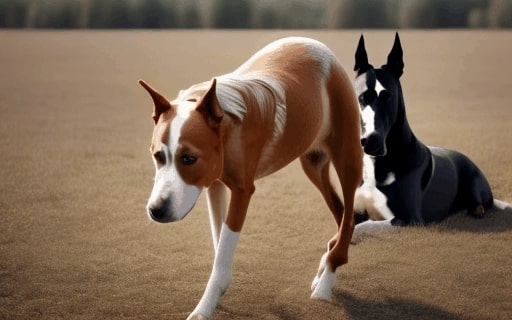} &
        \includegraphics[width=\linewidth]{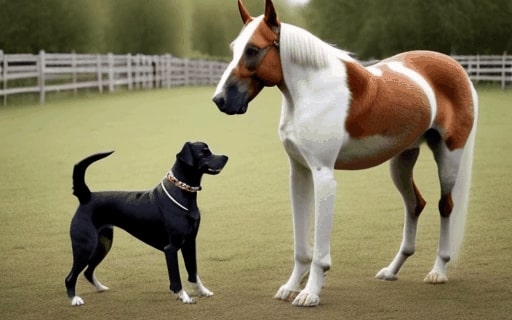}
    \end{tabular}

    \begin{tabularx}{\textwidth}{XX}
        \centering \footnotesize {\fontfamily{put}\selectfont "A person is strumming guitar"} & 
        \centering \footnotesize {\fontfamily{put}\selectfont "A dog and a horse"}
    \end{tabularx}

    \vspace{0.5em}
    \begin{tabularx}{\textwidth}{XXXX}
        \centering \small \textbf{Baseline} & 
        \centering \small \textbf{Free$^2$Guide} & 
        \centering \small \textbf{Baseline} &
        \centering \small \textbf{Free$^2$Guide}
    \end{tabularx}

    \renewcommand{\numColumns}{4}
    \renewcommand{\columnSpacing}{0.25em}

    \begin{tabular}{
        @{}
        p{\dimexpr(\textwidth-\columnSpacing*(\numColumns-1))/\numColumns} @{\hspace{\columnSpacing}}
        p{\dimexpr(\textwidth-\columnSpacing*(\numColumns-1))/\numColumns} @{\hspace{\columnSpacing}}
        p{\dimexpr(\textwidth-\columnSpacing*(\numColumns-1))/\numColumns} @{\hspace{\columnSpacing}}
        p{\dimexpr(\textwidth-\columnSpacing*(\numColumns-1))/\numColumns} @{}
    }
        \includegraphics[width=\linewidth]{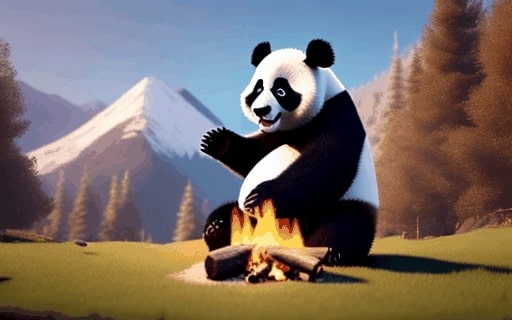} &
        \includegraphics[width=\linewidth]{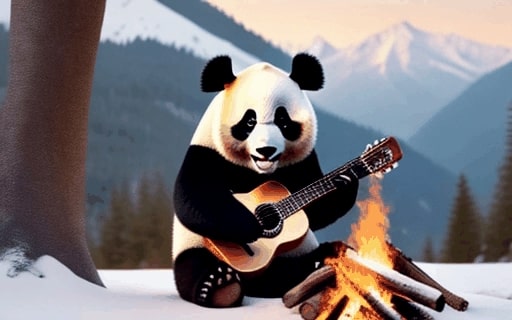} &
        \includegraphics[width=\linewidth]{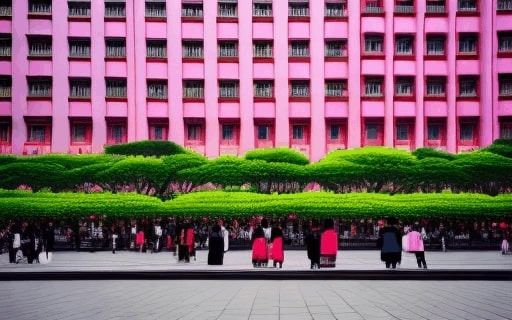} &
        \includegraphics[width=\linewidth]{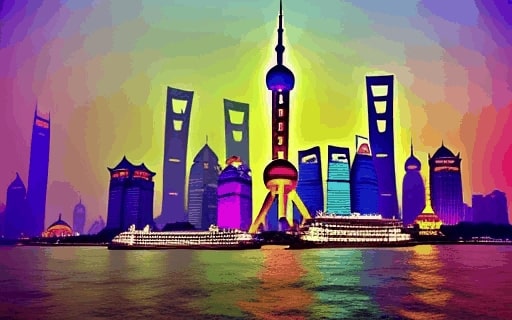}
    \end{tabular}

    \begin{tabularx}{\textwidth}{XX}
        \centering \footnotesize {\fontfamily{put}\selectfont "A happy fuzzy panda playing guitar nearby a campfire, snow mountain in the background"} & 
        \centering \footnotesize {\fontfamily{put}\selectfont "The bund Shanghai, vibrant color"}
    \end{tabularx}

    \vspace*{0.2cm}
    \captionof{figure}{
    Representative video results using \textbf{Free$^2$Guide}, a novel framework that enables training-\textbf{Free}, gradient-\textbf{Free} video \textbf{Guid}ance leveraging a Large Vision-Language Model. \textit{Each image shows the first frame of a video.}}
    \label{fig:main}
\end{center}
}]

\begin{abstract} 
Diffusion models have achieved impressive results in generative tasks for text-to-video (T2V) synthesis. However, achieving accurate text alignment in T2V generation remains challenging due to the complex temporal dependencies across frames. Existing reinforcement learning (RL)-based approaches to enhance text alignment often require differentiable reward functions trained for videos, hindering their scalability and applicability. In this paper, we propose \textbf{Free$^2$Guide}, a novel gradient-free  and training-free framework for aligning generated videos with text prompts. Specifically, leveraging principles from path integral control, Free$^2$Guide approximates guidance for diffusion models using non-differentiable reward functions, thereby enabling the integration of powerful black-box Large Vision-Language Models (LVLMs) as reward models. To enable image-trained LVLMs to assess text-to-video alignment, we leverage \textit{stitching} between video frames and use system prompts to capture sequential attributions. Our framework supports the flexible ensembling of multiple reward models to synergistically enhance alignment without significant computational overhead. Experimental results confirm that Free$^2$Guide using image-trained LVLMs significantly improves text-to-video alignment, thereby enhancing the overall video  quality. Our results and code are available at \href{https://kjm981995.github.io/free2guide/}{our project page} \footnote{\url{https://kjm981995.github.io/free2guide/}}.
\end{abstract}

\vspace{-0.4cm}

\section{Introduction}
\label{sec:Introduction}
Diffusion models~\cite{song2021scorebased, sohl2015deep, karras2022elucidating, rombach2022high} have emerged as powerful and versatile tools for generative modeling, achieving state-of-the-art results in tasks that require fine-grained control over content generation, such as text-to-image (T2I)~\cite{rombach2022high} and text-to-video (T2V) generation~\cite{ho2020denoising, dhariwal2021diffusion}. However, achieving perfect alignment with text conditions remains a significant challenge~\cite{gokhale2022benchmarking}. This issue becomes even more challenging in the video domain, where maintaining text-relevant content across frames requires handling complex temporal dependencies, often resulting in misalignment between generated frames and the given text prompt.

In the image domain, reinforcement learning (RL)-based methods have been introduced to address challenges in text-guided T2I generation by using reward models to estimate human preferences within diffusion models~\cite{xu2024imagereward, wu2023human, black2023training, fan2024reinforcement}. Previous works mainly focus on either directly fine-tuning the diffusion model with gradients derived from a reward function ~\cite{clark2023directly, prabhudesai2023aligning, prabhudesai2024video} or employing an RL-based policy gradient approach~\cite{black2023training,fan2024reinforcement}. While these fine-tuning methods can effectively improve sample alignment, they have notable limitations: the former requires a differentiable reward function, while the latter is typically limited to only few prompts. 

Directly adapting these text alignment approaches for the video domain presents two main challenges. First, they often require a dedicated video-specific reward function or additional training on curated video datasets. Collecting large-scale, aligned text-video datasets is far more complex than gathering image data, and developing reward functions tailored to video tasks is similarly difficult. Second, even with trained reward models for the video domain, additional challenges such as substantial memory demands for backpropagation emerge, which grow proportionally as model scale increases (\textit{i.e.}, scaling laws)~\cite{kaplan2020scaling}.

An alternative approach involves using differential reward models during inference time to guide diffusion models without fine-tuning model parameters~\cite{wallace2023end}. 
However, guidance-based methods still require a differentiable reward function, which excludes non-differentiable options like state-of-the-art visual-language model APIs or human preference-based metrics. To address this, recent studies have explored stochastic optimization to guide diffusion models during the sampling process using non-differentiable objective functions in music generation~\cite{huang2024symbolic}, and concurrent research extends this idea within the image domain~\cite{yeh2024training, zheng2024ensemble}. However, such methods cannot be directly applied to video diffusion models due to the complex temporal dependencies involved. 

To address these issues, here we introduce \textbf{Free$^2$Guide}---a novel text-to-video alignment method  by leveraging the temporal understanding capabilities of Large Vision-Language Models (LVLMs). 
Specifically, Free$^2$Guide  aligns text prompts in video generation without requiring gradients from the reward function. More specifically, drawing on principles from path integral control, Free$^2$Guide approximates guidance to align generated videos with text prompts, regardless of the reward function's differentiability. 
Another important contribution of this paper is a technique to adapt image-based LVLMs for temporal understanding.
In particular, we concatenate video frames in a structured grid layout, and design prompts that explicitly indicate sequence order and reasoning to help LVLMs evaluate videos more comprehensively.
By doing so, Free$^2$Guide enables the use of powerful black-box vision-language models as reward models, improving text-video alignment, as illustrated in Fig. \ref{fig:main}. Finally, our framework allows for the flexible combination of reward models by eliminating the need for computationally intensive fine-tuning and backpropagation. 
As such, we explore several combinatorial approaches to collaborate LVLMs with existing large-scale image-based models. Extensive experiments show that our methods improve text alignment and the quality of generated videos.

Our contributions are summarized as follows:
\begin{itemize}
    \item We introduce \textbf{Free$^2$Guide}, a novel framework for aligning generated videos with text prompts without requiring gradients from the reward function. To the best of our knowledge, Free$^2$Guide is the first gradient-free guidance approach for text-to-video generation that requires no additional training. 
    \item We adapt non-differentiable image-based LVLM APIs to enhance text-video alignment {by leveraging stitching and prompt design to capture video-specific attributes.} 
    \item We develop an effective ensemble approach that integrates large-scale image-based models to improve video generation guidance.
\end{itemize}

\section{Related Work}
\label{sec:RelatedWork}

\begin{figure*}[!t]
    \centering
    \includegraphics[width=\linewidth]{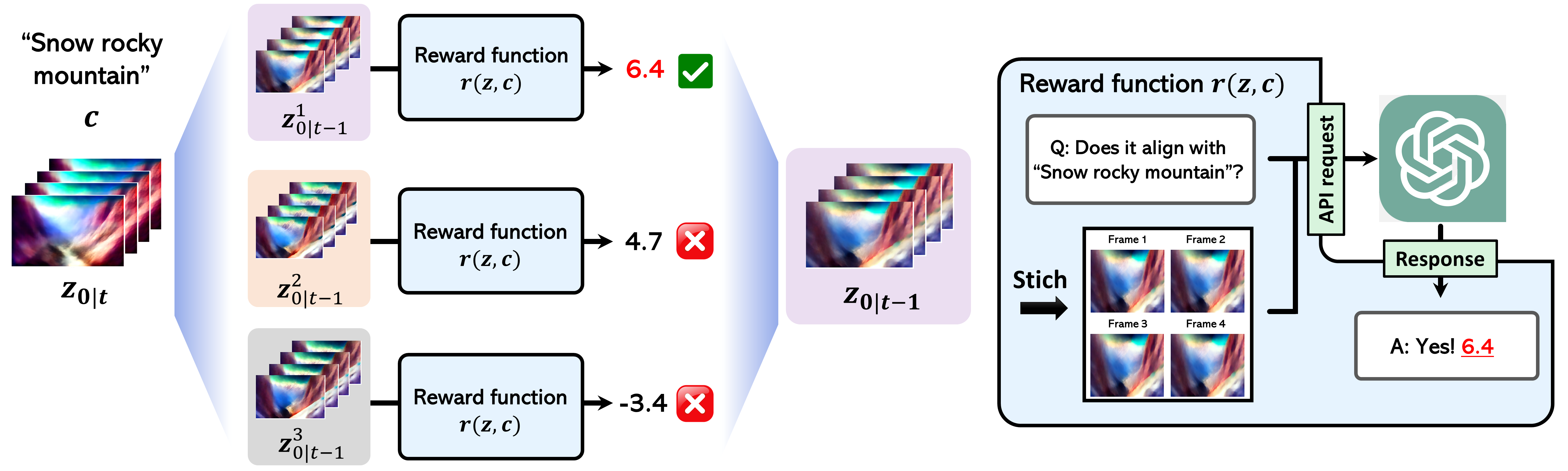}
    \caption{Overall pipeline of training-free gradient-free \textbf{Free$^2$Guide}. Free$^2$Guide leverages LVLMs' ability to comprehend stitched images, utilizing this capability to enhance frame-to-frame dynamic understanding and applying it within the video domain to improve text-video alignment. It also enables
 an effective ensemble approach that integrates large-scale image-based models to improve video generation guidance. 
    }
    \label{fig:pipeline}
    \vspace{-0.4cm}
\end{figure*}

\paragraph{Text-to-Video diffusion model}
Text-to-Video diffusion models (e.g., LaVie~\cite{wang2023lavie}, VideoCrafter~\cite{chen2023videocrafter1, chen2024videocrafter2}) employ diffusion processes to generate coherent video sequences from textual prompts~\cite{luo2023latent, he2022latent, ho2022video}. However, a notable limitation is that video diffusion models often struggle to generate videos that align accurately with the given text prompts, specifically in terms of spatial relationships (e.g., ``A on B'') and the representation of temporal style (e.g., ``zooming in'').

\noindent\textbf{Diffusion model with LVLM feedback}
While several approaches have been proposed to improve the diffusion generation process with Large Language Models (LLMs)~\cite{lian2023llm, wu2024self,feng2024layoutgpt, zhong2023adapter}, there has been limited exploration of methods leveraging Large Vision Language Models (LVLMs) that can also handle image domains. Recent works explore the integration of LVLMs as a feedback mechanism to image diffusion models to enhance control and guide diffusion processes. For instance, RPG~\cite{yang2024mastering} utilizes an LVLM as a planner to manipulate cross-attention layers in the diffusion model, while Demon~\cite{yeh2024training} demonstrates that LVLMs can guide diffusion in alignment with a given persona. In contrast, our approach leverages LVLMs' ability to comprehend stitched images, utilizing this capability to enhance frame-to-frame dynamic understanding and applying it within the video domain to improve text-video alignment.

\noindent\textbf{Human Preference Alignment via Reward Models}
Aligning with human preferences has improved generative quality in diffusion models through fine-tuning diffusion model using reward model gradients (DRaFT~\cite{clark2023directly}, AlignProp~\cite{prabhudesai2023aligning}) or policy gradients (DDPO~\cite{black2023training}, DPOK~\cite{fan2024reinforcement}). On the other hand, DOODL~\cite{wallace2023end} and Demon~\cite{yeh2024training} guide the denoising process to achieve text alignment without training diffusion models. Note, however, that the previously mentioned methods all focus on the image domain. Recent work VADER~\cite{prabhudesai2024video} fine-tunes a pre-trained video diffusion model using gradients of reward models for aesthetic and text-aligned generation. While this approach shows promising results for using video reward models, it demands substantial memory and does not utilize LVLMs. We address these limitations by proposing a text-video alignment method that approximates image reward gradients without fine-tuning.

\noindent\textbf{Zeroth-order gradient approximation}
Zeroth-order gradients, or gradient-free approaches, approximate gradients of non-differentiable functions by evaluating multiple points~\cite{liu2020primer, nesterov2017random}. In diffusion-based inverse problems, methods like EnKF~\cite{zheng2024ensemble} and SCG~\cite{huang2024symbolic} leverage gradient-free approximations to guide sampling based on non-differentiable or black-box forward models. However, there is a lack of research specifically focused on gradient-free approaches to guide sampling for video diffusion models. In video diffusion models, approximating a black-box reward model with a zeroth-order gradient is advantageous, as gradients of the reward are unavailable and the high-dimensional space of video data imposes memory limitations.

\section{Preliminaries}
\subsection{Video Latent Diffusion Model} 
Video Latent Diffusion Models (VLDMs) learn a stochastic process by iteratively denoising random noise generated by the forward diffusion process~\cite{dhariwal2021diffusion}
\begin{equation}
\label{eq:forward}
    q(\z_t | \z_0) = \mathcal{N}(\z_t; \sqrt{1 - \Bar{\alpha}_t} \, \z_{0}, \Bar{\alpha}_t \mathbf{I}),
\end{equation}
where $\z_0 = \mathcal{E}(\x)$ is the latent encoding of the clean video with encoder $\mathcal{E}$ and $\Bar{\alpha}_t$ is a noise scheduling coefficient at timestep $t$. The VLDM estimates the noise in $\z_t$ by minimizing the following objective:
\begin{equation}
\label{eq:denosing_objective}
    \mathbb{E}_{\z_0, \bm{\epsilon}, t, \vc} \left[ \|\bm{\epsilon} - \bm{\epsilon}_\theta(\z_t, t, \vc)\|^2 \right],
\end{equation}
where $\bm{\epsilon} \sim \mathcal{N}(0, \mathbf{I})$ and $\vc$ represents the conditioning input. 

To retrieve a clean latent representation, we use a reverse-time Stochastic Differential Equation (SDE) sampling process:
\begin{align}
\begin{split}
\label{eq:reverseSDE}
    d\z_t &= \Bar{\f}(\z_t) dt +  g(\z_t) \, d\bar{\mathbf{w}} \\
    &= \left[ \f(\z_t) - g(\z_t)^2 \nabla_{\z_t} \log p(\z_t) \right] dt + g(\z_t) \, d\bar{\mathbf{w}},
\end{split}
\end{align}
where $\f$ and $\Bar{\f}$ are the drift term for the forward SDE and reverse SDE, respectively, $g$ is the diffusion coefficient, and $\bar{\mathbf{w}}$ represents a reverse time Wiener process. The initial point for reverse SDE is sampled from a normal Gaussian distribution. By discretizing the reverse SDE with an appropriate noise schedule, the VLDM retrieves a clean latent representation based on the DDIM~\cite{song2020denoising} trajectory, 
\begin{align}
\begin{split}
\label{eq:reverse_sampling-2}
    \sigma_t &:= \eta \sqrt{\left(\frac{1 - \Bar{\alpha}_{t-1}}{1 - \Bar{\alpha}_t}\right) \left(1 - \frac{\Bar{\alpha}_t}{\Bar{\alpha}_{t-1}}\right)} \\
    \z_{0|t} &= \frac{1}{\sqrt{\Bar{\alpha}_t}} \left( \z_t - \sqrt{1 - \Bar{\alpha}_t} \epsilon_\theta(\z_t, t, \vc) \right) \\
    \z_{t-1} &= \sqrt{\Bar{\alpha}_{t-1}} \z_{0|t}  + \sqrt{1 - \Bar{\alpha}_{t-1} - \sigma_t^2} \bm{\epsilon}_\theta(\z_t, t, \vc) + \sigma_t\epsilonb,    
\end{split}
\end{align}
where $\sigma_t$ controls the stochasticity of sampling, $\epsilonb \sim \Nc(0, \rmI)$ and $\z_{0|t} = \Ed [\z_0|\z_t]$ denotes the posterior mean or denoised version of $\z_t$, computed by Tweedie's formula~\cite{efron2011tweedie}. To transform the latent representation back to the video domain, a decoder $\mathcal{D}$ is used to decode the latent.

\subsection{Guidance in Diffusion Model}
Given the reverse SDE in \eqref{eq:reverseSDE}, our goal is to obtain the optimal control $\vu(\z_t)$ :
\begin{align}
\label{eq:optimal_reverseSDE}
    d\z_t &= \left[ \Bar{f}(\z_t) + \vu(\z_t)\right] dt + g(\z_t) \, d\bar{\mathbf{w}},
\end{align}
which directs the sampling process toward target distribution $p(\z_t|y)$, where $y$ represent a desired condition, such as label, class or text prompt~\cite{williams1979diffusions}. 
In classifier guidance~\cite{nie2022diffusion}, if an auxiliary classifier is available to estimate the likelihood $p(y|\z_t)$, the control term can be defined as 
\begin{equation}
\label{eq:cf-guidance}
 \vu(\z_t) = -g(\z_t)^2 w \nabla_{\z_t} \log p(y|\z_t),
\end{equation}
where $w$ is a scaling factor that adjusts the strength of the guidance. This control term follows from applying the Bayes rule to express $p(\z_t|y) \propto p(\z_t|y)p(y|\z_t)^w$.

One might consider adapting classifier guidance by treating the reward model as a classifier. However, this approach presents two challenges: the reward model is not trained on noisy latent representations $\z_t$ and requires differentiability. To alleviate these limitations, we utilize a path integral control approach with zeroth-order gradient approximation, as described in the following Section \ref{sec.pic}.

\subsection{Path Integral Control}
\label{sec.pic}
Considering the diffusion model as an entropy regularized Markov Decision Process (MDP), we can conceptualize reverse SDE in the Reinforcement Learning (RL) framework~\cite{uehara2024understanding, black2023training, fan2024reinforcement} with the state $\vs_t$ and the action $\va_t$ corresponding to the input $\z_t$. In this formula, the optimal policy $p^*$ maximizes the following objective:
\begin{equation}
\label{eq:MDP}
    \Ed_{p} [\vr(\z_0) - \alpha \sum_{\tau=T}^1 D_{KL}(p(\z_{\tau-1}|\z_\tau) || p_\theta(\z_{\tau-1}| \z_\tau ))],
\end{equation}
where $\alpha$ is a coefficient of KL divergence with original policy $p_\theta$ defined by diffusion model. Let $p_\theta(\z_{t-1}|\z_t) = \Nc(\mub_t, \sigma_t^2 \mI)$ be a reverse transition distribution in the SDE for the diffusion model and $p_\theta(\z_{0:t}) := p_\theta(\z_t) \Pi_{\tau=1}^t p(\z_{\tau-1}|\z_\tau)$. We can define a value function as 
\begin{align}
\label{eq:prop1-2}
\begin{split}
   \exp{\left(\frac{\vv(\z_t)}{\alpha}\right)} &= \int \exp{\left(\frac{\vv(\z_{t-1})}{\alpha}\right)} p_\theta(\z_{t-1}|\z_t) d\z_{t-1} \\ &=
   \Ed_{p_\theta(\z_{0:t})} \left[ \exp \left( \frac{\vr(\z_0)}{\alpha} \right) | \z_t \right],
\end{split}
\end{align}
satisfying $\vv(\z_0)=\vr(\z_0)$ is a reward function~\cite{uehara2024understanding}.

The optimal control to address the entropy-regularized MDP system can be obtained by solving the Hamilton-Jacobi-Bellman (HJB) equation as follows~\cite{uehara2024fine, huang2024symbolic}:
\begin{equation}
\label{eq:hjb}
    \vu(\z_t) = -\frac{\sigma_t^2 \nabla_{\z_t} \vv(\z_t)} {\alpha}.
\end{equation}
However, this term requires the gradient of the value function. To bypass the gradient requirements, one can use path integral control, which is an approach to estimate the optimal control (or guidance) based on the principles of stochastic optimal control~\cite{theodorou2010generalized, kappen2005path, uehara2024fine}. 
In \cite{huang2024symbolic}, the optimal control can be approximated as
\begin{align}
\label{eq:approx.grad}
\begin{split}
    \vu(\z_t) &\simeq  -\frac{\Ed\left[\exp\left( \frac{\vr(\z_{0})}{\alpha}\right) (\z_{t-1} - \mub_t)| \z_t \right]}{\Ed\left[\exp\left( \frac{\vr(\z_{0})}{\alpha} \right) | \z_t \right]}.
\end{split}
\end{align}
While SCG~\cite{huang2024symbolic} utilizes this optimal control with diffusion models to solve inverse problems in image domain, we aim to use LVLMs to guide videos toward improved text alignment.

\section{Free$^2$Guide}
\label{sec:Method}

In this section, we introduce Free$^2$Guide, a framework that uses a non-differentiable reward model to guide video generation during the sampling process. In Sec. \ref{sec4}, we discuss how to apply image-based reward models, including LVLM, for text-video alignment. Sec. \ref{sec3} outlines methods for ensembling multiple reward models to achieve synergistic effects. Finally, we interpret the diffusion model as an entropy-regularized MDP and describe its practical implementation (Sec. \ref{sec2}).

\subsection{Video Guidance leveraging Image LVLMs}
\label{sec4}

\noindent\textbf{Motivation}
By leveraging the path integral control approach discussed in Sec. \ref{sec.pic}, we can guide the reverse process without relying on the gradient of the reward function. If the reward model $\vr$ in \eqref{eq:approx.grad} assesses the alignment of the generated video with the text prompt, it can help steer the video output to enhance fidelity to the prompt. However, due to the complexity of videos compared to static images, there are limited large-scale models specifically trained for video and text alignment. {We analyze the impact of video-based reward models on video guidance and find that their effectiveness is limited (see Appendix~\ref{app.D}).} 

Applying these image-based reward models directly for video guidance, of course, presents challenges. Image-based models are not designed to process time-dependent features, such as motion, flow, and dynamics, so specific adaptations are required for these models to assess text-video alignment. As shown in Algorithm \ref{alg:reward}, we calculate the reward for a video by summing frame-by-frame rewards from the image-based model. This approach enables alignment with spatial information within individual video frames but still lacks guidance on temporal dynamics.

\paragraph{Image-based LVLMs as a Video Reward Model} Although LVLMs are trained on static image-text data, their extensive pretraining across diverse visual contexts enables them to implicitly capture elements of motion. {As shown in Table 1 of \cite{sun2024t2v}, treating video as an image grid in LVLMs strongly correlates with human evaluation. Furthermore, results from \cite{li2024mvbench, kim2024image} demonstrate that image-based LVLMs achieve performance comparable to video-specific LLMs in video QA, validating our approach. 

Accordingly, to adapt LVLMs for evaluating multiple frames simultaneously, we employ a method called \textit{stitching}, which combines key frames into a single composite image (see Fig. \ref{fig:pipeline}). Specifically, we first sample key frames from the video and arrange them in a structured grid layout, labeling each frame with its index to indicate its position in the sequence. This approach allows LVLMs to process temporal information by leveraging spatial relationship between frames.} 

Then, to help LVLMs understand frame order within the composite image, we provide explicit sequence instructions through a system prompt. This efficient adaptation enables LVLMs to recognize frame order by referencing frame numbers rather than processing them linearly. {We incorporate Zero-shot Chain-of-Thought~\cite{kojima2022large} in the system prompt to enhance reasoning ability and mitigate hallucinations. In the user prompt, we instruct the LVLM to consider every key frame individually and evaluate the alignment score between the composite image and the text prompt on a scale of 1 to 9.} The full system instructions and query templates are detailed in Appendix~\ref{sec:implementation}.

\begin{figure}[!t]
\begin{minipage}{1.0\columnwidth}
    \begin{algorithm}[H]
    \setstretch{1.1}
        \caption{Reward Model $\vr(\mathcal{D}(\z_{0|t}), \vc)$}
        \label{alg:reward}
        \begin{algorithmic}[1]
            \Require Reward function $\vr$, condition $\vc$, prompt $\vp$, decoded frames $\x_{0|t} := \mathcal{D}(\z_{0|t})$, and key frames $k \subset [1,N]$
            \If{$\vr$ is CLIP}
                \State reward $\gets \sum_{i \in k} \texttt{sim} (\vr(\x_{0|t}^{i}), \vr(\vc))$
            \ElsIf{$\vr$ is ImageReward}
                \State reward $\gets \sum_{i \in k} \vr (\x_{0|t}^{i},\vc)$
            \ElsIf{$\vr$ is LVLM}
                \State reward $\gets \vr(\texttt{concat}_{i\in k}(\vx_{0|t}^{i}), \vc, \vp)$
            \EndIf
            \State {\bfseries return} reward
        \end{algorithmic}
    \end{algorithm}

    \vspace{-0.4cm}
    
    \begin{algorithm}[H]
    \setstretch{1.1}
    \caption{Free$^2$Guide}
    \label{alg:GFVG}
    \begin{algorithmic}[1]
        \Require Video diffusion model $\epsilonb_\theta$, reward function $\vr$, decoder $\mathcal{D}$, noise scheduling parameter $\{\Bar{\alpha}_t\}_{t=1}^T, \{\sigma_t\}_{t=1}^T$
        \For{$t=T$ {\bfseries to} $1$}
            \State $\z_{0|t} \gets \frac{1}{\sqrt{\Bar{\alpha}_{t-1}}} \left( \z_t - \sqrt{1 - \Bar{\alpha}_t} \epsilonb_\theta(\z_t) \right)$
            \State $\hat{\z}_{t-1} \gets \sqrt{\Bar{\alpha}_t}\z_{0|t} + \sqrt{1-\Bar{\alpha}_{t-1}-\sigma_t^2}\epsilonb_\theta(\z_t)$
            \State $\epsilonb^1, \cdots, \epsilonb^n \sim \Nc(0, \rmI)$
            \State $\z_{t-1}^i \gets \hat{\z}_{t-1} + \sigma_t \epsilonb^i$
            \State $\z_{0|t-1}^i \gets \frac{1}{\sqrt{\Bar{\alpha}_{t-1}}} \left( \z_{t-1}^i - \sqrt{1 - \Bar{\alpha}_{t-1}} \epsilonb_\theta(\z_{t-1}^i) \right)$

            \State $\vr_1 \gets $LVLM
            \If{Ensemble}
            \State $\vr_2 \in \{\text{CLIP, ImageReward}\}$
            \State $j \gets \text{argmax}_i$ Reward$_{\text{ens}}(\mathcal{D}(\z_{0|t-1}^i), \vr_1, \vr_2) \quad$ From Sec. \ref{sec3}.
            \Else
            \State $j \gets \text{argmax}_i$ $\vr_1(\mathcal{D}(\z_{0|t-1}^i), \vc)$ 
            \EndIf
            \State $\z_{t-1} \gets \z_{t-1}^j$
        \EndFor
        \State {\bfseries return} $\z_0$
    \end{algorithmic}
    \end{algorithm} 
\end{minipage}
\vspace{-0.4cm}
\end{figure}

\subsection{Ensembling Reward Functions}
\label{sec3}

Unlike gradient-based guidance, our method significantly reduces memory requirements by avoiding the computationally intensive backpropagation process. This enables us to concurrently employ multiple rewards for sampling guidance, potentially leading to synergistic benefits with large-scale image models. We explore ensemble methods that allow LVLMs to incorporate temporal information, thereby supporting more effective guidance for video alignment when combined with large-scale image models. Note that Demon~\cite{yeh2024training}, a concurrent work that also proposed ensemble rewards, failed to show the synergy effect of ensemble and did not have to handle temporal information.


Given the $n$ videos $\{V_i\}_{i=1}^n$, we propose three ensembling methods to combine multiple reward models: Weighted Sum, Normalized Sum, and Consensus.

\begin{itemize}
    \item \textbf{Weighted Sum}: 
    This method combines the outputs by computing a fixed weighted sum, allowing us to control the influence of each reward model.
    \begin{equation}
     \text{Reward}_{\text{ens}}(V_i, \vr_1, \vr_2) = \beta \vr_1(V_i) + (1 - \beta) \vr_2(V_i) ,  
    \end{equation}
    where $\beta \in [0, 1]$ is a constant weight factor that balances the contributions of reward models $\vr_1$ and $\vr_2$.

    \item \textbf{Normalized Sum}:
    To ensure a balanced contribution of each reward models, we first normalize each reward's output to the range $[0, 1]$, then sum these normalized values to get the final ensemble reward. 
    \begin{equation}
     \text{Reward}_{\text{ens}}(V_i, \vr_1, \vr_2) = \sum_{\vr} \frac{\vr(V_i) - \min(\vr(V_i))}{\max(\vr(V_i)) - \min(\vr(V_i))},
    \end{equation}
    where $\max(\vr), \min(\vr)$ represents the maximum and minimum score from $n$ reward outputs.

    \item \textbf{Consensus}:
    Inspired by the Borda count~\cite{emerson2013original}, each reward model ranks the videos from best to worst, assigning $\text{points}_{\vr}$ based on their rank. The top-ranked video receives the maximum points, down to 1 point for the lowest rank. The total reward for each video $V_i$ is the sum of points from both reward model.
    \begin{equation}
     \text{Reward}_{\text{ens}}(V_i, \vr_1, \vr_2) = \text{points}_{\vr_2}(V_i) + \text{points}_{\vr_1}(V_i).
    \end{equation}
\end{itemize}

\subsection{Guidance using Path Integral Control}
\label{sec2}
To guide the reverse sampling process without computing the gradient of the reward function, we utilize the framework outlined in \eqref{eq:approx.grad}. However, the expectation of the reward function in \eqref{eq:approx.grad}  demands extensive network function evaluations (NFE) by solving complex differential equations, such as PF-ODE~\cite{song2021scorebased}. Inspired by \citep{huang2024symbolic}, we instead apply the DPS~\cite{chung2023diffusion} approach to approximate \eqref{eq:prop1-2} by using the posterior mean of $\z_t$, as defined in \eqref{eq:reverse_sampling-2}. Following DPS, we can set $p(\z_{0:t}) = \delta(\z -\Ed \left[ \z_0 | \z_t\right])$ using Direc delta distribution $\delta$ in which case \eqref{eq:approx.grad} becomes:
\begin{equation}  
\label{eq:optimal_contol_dps}
    \vu(\z_t) \simeq - \frac{\Ed_{p_\theta(\z_{t-1}|\z_t)} \left[ \exp \left( \frac{\vr(\z_{0|t-1})}{\alpha}\right) (\z_{t-1} - \mub_t)\right]}{\exp\left( \frac{\vr(\z_{0|t})}{\alpha} \right)}.
\end{equation}
To approximate this expectation using the Monte Carlo method, we sample $n$ different $\z_{t-1}$ through the reverse SDE as outlined in \eqref{eq:reverse_sampling-2}. Then we  assume $\alpha \rightarrow 0$ to obtain optimal control. Under this assumption, \eqref{eq:reverseSDE} becomes equivalent to selecting the $\z_{t-1}$ that maximizes the reward of $\z_{0|t-1}$~\cite{huang2024symbolic}. While \cite{huang2024symbolic} arbitrarily weighted the reward function and assumed the weight to be zero, we interpret this as relaxing the entropy-regularization term in \eqref{eq:MDP} by defining the diffusion process as an entropy-regularized MDP. In practical terms, this approach eliminates careful parameter exploration by selecting $\z_{t-1}$ with the largest reward. 

By following this adjusted sampling strategy as described in Algorithm~\ref{alg:GFVG}, Free$^2$Guide can efficiently steer video generation towards better alignment with the reward signals.



\begin{figure*}[!t]
    \centering
    \includegraphics[width=\linewidth]{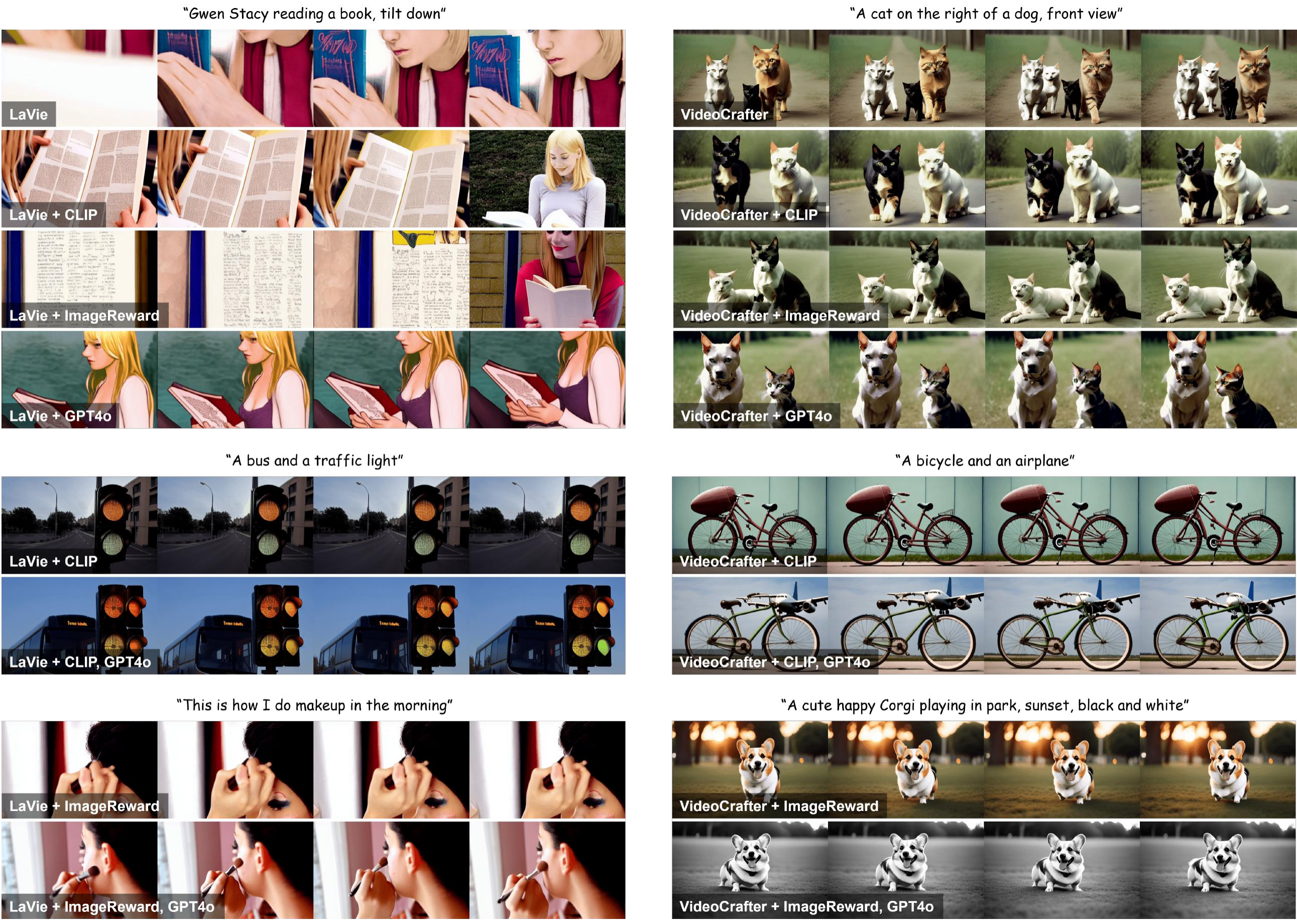}
    \vspace{-0.4cm}
    \caption{Qualitative results of our method. Comparison with LaVie on the left and VideoCrafter2 on the right.
    }
    \label{fig:results_t2v}
    \vspace{-0.2cm}
\end{figure*}


\section{Experiments}
\label{sec:Experiments}

\noindent\textbf{Baselines and Sampling Strategy.} 
We use open-source text-to-video diffusion models, LaVie~\cite{wang2023lavie} and VideoCrafter2~\cite{chen2024videocrafter2}, as baseline models. The generated videos contain 16 frames with a resolution of $320\times512$. We employ LVLM as \texttt{GPT-4o-2024-08-06}~\cite{achiam2023gpt} using OpenAI APIs. We employ two large-scale models CLIP~\cite{radford2021learning} and ImageReward~\cite{xu2024imagereward}, to validate that LVLM’s capability to account for temporal dynamics can enhance text-video alignment when used alongside large-scale image reward models. In CLIP, we can assess alignment by measuring cosine similarity between text and image embeddings. On the other hand, we can use ImageReward output as an reward since it predicts human preference for image-text pairs.  For adaptation to the video domain, we extract key frames from each denoised video and sum the reward for each frame to evaluate overall alignment, as outlined in  Algorithm~\ref{alg:reward}.

We employ stochastic DDIM sampling with $\eta = 1$ in \eqref{eq:reverse_sampling-2} for a total of $T = 50$ steps and apply classifier-free guidance~\cite{ho2022classifier} using a guidance scale of $w = 7.5$ for LaVie and $w = 12$ for VideoCrafter2. The number of samples at each guidance step is set to $n = 5$ for LaVie and $n = 10$ for VideoCrafter2. Guidance is applied during the early sampling steps, specifically within $t \in [T, T-5]$. In the weighted sum ensemble, we assign a weight of $\beta = 0.75$ to the LVLM reward.

\begin{table}[!b]
    \centering
    \begin{minipage}{0.45\linewidth}
        \centering
        \resizebox{1.0\textwidth}{!}{
        \begin{tabular}{lc}
            \toprule
            Method  & Avg. \\
            \midrule
            LaVie + CLIP & 0.5712 \\
            + GPT\textsubscript{Weighted Sum} & \textbf{0.5738} \\
            + GPT\textsubscript{Normalized Sum} & 0.5734 \\
            + GPT\textsubscript{Consensus} & 0.5679 \\
            \bottomrule
        \end{tabular}
        }
    \end{minipage}\hfill
    \begin{minipage}{0.5\linewidth}
        \centering
        \resizebox{1.0\textwidth}{!}{
        \begin{tabular}{lc}
            \toprule
            Method & Avg. \\
            \midrule
            LaVie + ImageReward & 0.5676 \\
            + GPT\textsubscript{Weighted Sum} & \textbf{0.5726} \\
            + GPT\textsubscript{Normalized Sum} & 0.5715 \\
            + GPT\textsubscript{Consensus} & 0.5692 \\
            \bottomrule
        \end{tabular}
        }
    \end{minipage}
    \caption{Qualitative comparison between ensemble methods. }
    \label{tab:ensemble}
    \vspace{-0.2cm}
\end{table}

\begin{table*}[t]
\centering
\renewcommand{\arraystretch}{1.1}
\resizebox{1.0\textwidth}{!}{
\begin{tabular}{lccccccc}
\toprule
 \multicolumn{1}{c}{\scalebox{1.2}{\textbf{Method}}} & {Appearance Style} & {Temporal Style} &   {Human Action} &   {Multiple Objects} &   {Spatial Relationship} &   {Overall Consistency} &  \multicolumn{1}{c}{\scalebox{1.2}{\textbf{Avg.}}}\\
\midrule
\rowcolor{LightCyan} 
   {\scalebox{1.2}{LaVie \cite{wang2023lavie}}} & 0.2312& 0.2502  & 0.9300  & 0.2027  & 0.3496 & 0.2694   & 0.3722 \\
      {\scalebox{1.2}{+ GPT}} & 0.2366 \red{\scalebox{0.8}{(+2.3\%)}}&0.2508	\red{\scalebox{0.8}{(+0.2\%)}}& 0.9300 \scalebox{0.8}{(-0.0\%)}& 0.2546 \red{\scalebox{0.8}{(+25.6\%)}}&0.3531	\red{\scalebox{0.8}{(+1.0\%)}}&0.2709 \red{\scalebox{0.8}{(+0.6\%)}}&0.3827  \\
    \midrule
   {\scalebox{1.2}{+ CLIP}} & 0.2370 \red{\scalebox{0.8}{(+2.5\%)}}& 0.2490 \blue{\scalebox{0.8}{(-0.5\%)}} & 0.9400 \red{\scalebox{0.8}{(+1.1\%)}}& 0.2607 \red{\scalebox{0.8}{(+28.6\%)}}& 0.3074 \blue{\scalebox{0.8}{(-12.1\%)}}& 0.2738 \red{\scalebox{0.8}{(+1.6\%)}}& 0.3780\\
 {\scalebox{1.2}{++ GPT}} & 0.2350 \red{\scalebox{0.8}{(+1.6\%)}}&0.2487	\blue{\scalebox{0.8}{(-0.6\%)}}& 1.000 \red{\scalebox{0.8}{(+7.5\%)}}& 0.2447 \red{\scalebox{0.8}{(+20.7\%)}}&0.3180	\blue{\scalebox{0.8}{(-9.0\%)}}&0.2742 \red{\scalebox{0.8}{(+1.7\%)}} & \textbf{0.3868}\\
 \midrule
   {\scalebox{1.2}{+ ImageReward}} &0.2360 \red{\scalebox{0.8}{(+2.1\%)}}&0.2483 \blue{\scalebox{0.8}{(-0.8\%)}}&0.9300 \scalebox{0.8}{(-0.0\%)}&0.2637 \red{\scalebox{0.8}{(+30.1\%)}}&0.2614 \blue{\scalebox{0.8}{(-25.2\%)}}&0.2728 \red{\scalebox{0.8}{(+1.2\%)}}&0.3687  \\
{\scalebox{1.2}{++ GPT}} & 0.2373 \red{\scalebox{0.8}{(+2.6\%)}}&0.2497	\blue{\scalebox{0.8}{(-0.2\%)}}& 0.9400 \red{\scalebox{0.8}{(+1.1\%)}}& 0.2462 \red{\scalebox{0.8}{(+21.4\%)}}&0.3014	\blue{\scalebox{0.8}{(-13.8\%)}}&0.2772 \red{\scalebox{0.8}{(+2.9\%)}} & 0.3753\\
\midrule

\midrule
\rowcolor{LightCyan} 
   {\scalebox{1.2}{VideoCrafter2 \cite{chen2024videocrafter2}}} & 0.2490&0.2567 &	0.9300	&0.3880 &0.3760 	&0.2778		&0.4129 \\
   {\scalebox{1.2}{+ GPT}} &0.2504 \red{\scalebox{0.8}{(+0.6\%)}}	&0.2568 \red{\scalebox{0.8}{(+0.0\%)}}	&0.9500 \red{\scalebox{0.8}{(+2.2\%)}}	&0.4878 \red{\scalebox{0.8}{(+25.7\%)}}&0.4225 \red{\scalebox{0.8}{(+12.4\%)}}	&0.2872 \red{\scalebox{0.8}{(+3.4\%)}}		&0.4425 \\

\midrule
   {\scalebox{1.2}{+ CLIP}}& 0.2542 \red{\scalebox{0.8}{(+2.1\%)}}&0.2621	\red{\scalebox{0.8}{(+2.1\%)}}& 0.9300 {\scalebox{0.8}{(-0.0\%)}}&0.4261 \red{\scalebox{0.8}{(+9.8\%)}}	&0.2923 \blue{\scalebox{0.8}{(-22.3\%)}}&0.2802 \red{\scalebox{0.8}{(+0.9\%)}}		&0.4075  \\
         {\scalebox{1.2}{++ GPT}} & 0.2490 {\scalebox{0.8}{(+0.0\%)}}&0.2612	\red{\scalebox{0.8}{(+1.8\%)}}& 0.9600 \red{\scalebox{0.8}{(+3.2\%)}}& 0.4474 \red{\scalebox{0.8}{(+15.3\%)}}&0.3361	\blue{\scalebox{0.8}{(-10.6\%)}}&0.2837 \red{\scalebox{0.8}{(+2.1\%)}} & 0.4229 \\
         \midrule
   {\scalebox{1.2}{+ ImageReward}} & 0.2513	\red{\scalebox{0.8}{(+0.9\%)}}&0.2574 \red{\scalebox{0.8}{(+0.3\%)}} &0.9700 \red{\scalebox{0.8}{(+4.3\%)}}	&0.4733 \red{\scalebox{0.8}{(+22.0\%)}}	&0.4264 \red{\scalebox{0.8}{(+13.4\%)}}&0.2826 \red{\scalebox{0.8}{(+1.7\%)}}			&0.4435 \\
            {\scalebox{1.2}{++ GPT}} & 0.2533 \red{\scalebox{0.8}{(+1.7\%)}}&0.2607	\red{\scalebox{0.8}{(+1.6\%)}}& 0.9400 \red{\scalebox{0.8}{(+1.1\%)}}& 0.5160 \red{\scalebox{0.8}{(+33.0\%)}}&0.4371	\red{\scalebox{0.8}{(+16.3\%)}}&0.2828 \red{\scalebox{0.8}{(+1.8\%)}} & \textbf{0.4483} \\
\bottomrule
\end{tabular}
}
\vspace{-0.1cm}
\caption{
Quantitative evaluation on text alignment. Higher numbers indicate better alignment with the text prompt. The numbers in parentheses denote the performance difference from the baseline.
}
\vspace{-0.3cm}
\label{tab:results_t2v}
\end{table*}

\begin{table}[t]
\centering
\renewcommand{\arraystretch}{1.1}
\resizebox{1.0\linewidth}{!}{
\Large
\begin{tabular}{lccccccc}
\toprule
 \multicolumn{1}{c}{\scalebox{1.2}{\textbf{Method}}} & \makecell{Subject\\Consistency} & \makecell{Background\\Consistency} &   \makecell{Motion\\Smoothness} &   \makecell{Dynamic\\Degree} &   \makecell{Aesthetic\\Quality} &   \makecell{Imaging\\Quality} &  \multicolumn{1}{c}{\scalebox{1.2}{\textbf{Avg.}}}\\
\midrule
\rowcolor{LightCyan} 
   {\scalebox{1.2}{LaVie \cite{wang2023lavie}}} & 0.9450& 0.9689  & 0.9718  & 0.4799  & 0.5687 & 0.6611 &0.7659   \\
      {\scalebox{1.2}{+ GPT}} & 0.9470  &0.9693	 & 0.9742  & 0.4725  &0.5726	 &0.6615   & 0.7662  \\
      \midrule
   {\scalebox{1.2}{+ CLIP}} & 0.9495  & 0.9712   & 0.9735  & 0.4560  & 0.5727  & 0.6637   & 0.7644\\
     {\scalebox{1.2}{++ GPT}} & 0.9622  &0.9781	 & 0.9804  & 0.3703  &0.5951	 &0.6795   & 0.7609 \\
     \midrule
   {\scalebox{1.2}{+ IR}} &0.9443  &0.9681  &0.9732  &0.4872  &0.5664  &0.6605   & 0.7666  \\
        {\scalebox{1.2}{++ GPT}} & 0.9758  &0.9813	 & 0.9832  & 0.5165  &0.5662	 &0.6530   & \textbf{0.7699} \\
    \midrule

\midrule
\rowcolor{LightCyan} 
   {\scalebox{1.2}{VC2 \cite{chen2024videocrafter2}}} & 0.9658&0.9748 &	0.9818	&0.3846 &0.5860 	&0.6772 & \textbf{0.7617}  \\
               
   {\scalebox{1.2}{+ GPT}} &0.9746  	&0.9800  	&0.9827  	&0.2949  &0.5977  	&0.6924  	& 0.7537 \\
   \midrule
   {\scalebox{1.2}{+ CLIP}}& 0.9762  &0.9816	 & 0.9839  &0.2491  	&0.6037  &0.6886   & 0.7472\\
         {\scalebox{1.2}{++ GPT}} & 0.9770  &0.9823	 & 0.9838  & 0.2399  &0.6042	 &0.6878   & 0.7458 \\
\midrule
   {\scalebox{1.2}{+ IR}} & 0.9739	 &0.9801   &0.9828  	&0.2711  	&0.5994  &0.6857   & 0.7488			 \\
            {\scalebox{1.2}{++ GPT}} & 0.9758  &0.9813	 & 0.9832  & 0.2564  &0.6039	 &0.6877   & 0.7480 \\
\bottomrule
\end{tabular}
}
\vspace{-0.1cm}
\caption{
Comparison of the general quality of the generated video independent of the text prompt. Higher numbers indicate better video quality. `VC2' is VideoCrafter2 and `IR' is ImageReward.
}
\vspace{-0.4cm}
\label{tab:results_t2v2}
\end{table}

\noindent\textbf{Text Alignment Evaluation.}
We conduct quantitative evaluation using VBench~\cite{huang2023vbench}, a benchmark designed to evaluate the alignment of text-to-video (T2V) models with respect to a text prompt. Our evaluation protocol measures text alignment across six dimensions: Appearance Style, Temporal Style, Human Action, Multiple Objects, Spatial Relationship and Overall Consistency. For a fair comparison, we use standardized prompts for each metric, ensuring consistent conditions across different models. 


\noindent\textbf{General Video Quality Evaluation.} In addition to text alignment, we evaluate the general quality of generated videos independently of text prompts using six metrics in VBench: Subject Consistency, Background Consistency, Motion Smoothness, Dynamic Degree, Aesthetic Quality, and Imaging Quality. 

\noindent\textbf{Video-specific Attributes.}
Since VBench prompts involve limited movement, we conducted additional experiments using T2V-CompBench~\cite{sun2024t2v} to analyze video-specific motion and dynamics. We measure Dynamic Attribution Binding, which evaluates how well models handle state changes (\textit{e.g.} shape and texture) and color variations over time.

\subsection{Results}
In this section, we present both qualitative and quantitative results to demonstrate the effectiveness of our method. The top four rows of Fig. \ref{fig:results_t2v} shows visual comparisons between the baseline and reward models. We observe that leveraging the GPT-4o model to assess text-video alignment improves alignment with respect to temporal dynamics (\eg "tilt down") and semantic representation (\eg "A and B"). These results indicate that LVLM can account for temporal information by processing multiple sub-frames of video simultaneously, with strong performance in spatial understanding. 

Building on LVLMs' capability to account for temporal dynamics, we validate the feasibility of ensembling techniques that integrate guidance from large-scale image models to improve text-video alignment. This approach enables LVLMs to process temporal information, enhancing the quality of guidance. In Table \ref{tab:ensemble}, we explore the most effective ensemble method by comparing average scores on text alignment and general video quality evaluation from VBench. We find that assigning more weight to LVLM outperformed the alternative of balancing model contributions equally in the ensemble, indicating that the role of LVLM is significant. Thus, we adopt the weighted sum ensemble as the default setting. The bottom four rows of Fig. \ref{fig:results_t2v} also illustrate qualitative results for ensembling, showing that combining GPT-4o with other image reward models accurately resolves issues related to dynamics or multiple objects that standalone reward models struggle to properly identify, while maintaining overall structure.


For more detailed evaluations, we compare the quantitative results in Table \ref{tab:results_t2v} to assess text-video alignment. Analysis of the average evaluation scores reveals that incorporating LVLM consistently outperforms configurations that exclude it. Specifically, we observe the most significant improvement in handling Spatial Relationship across baselines. Since CLIP has a limited zero-shot spatial reasoning capability~\cite{subramanian2022reclip}, the text alignment performance decreases in Spatial Relationship when using CLIP alone. However, ensembling with LVLM offers additional cues that help CLIP to better account for spatial semantics, leading to performance improvements. Furthermore, incorporating LVLM enhances Human action, Overall Consistency in overall case and Temporal Style, except when using CLIP as the reward model. Since LVLM can understand temporal nuances by processing multiple frames at once, it improves performance by supporting the alignment of temporal movement.

Additionally, we compare general video quality in Table \ref{tab:results_t2v2}. We confirm that even without explicit guidance for consistency or motion, alignment with text prompts improves most quality metrics except for Dynamic Degree. This metric often trades off with consistency but can be improved by ensembling GPT-4o with ImageReward in the LaVie model. This suggests that ImageReward compensates for the performance drop in Dynamic Degree that GPT-4o alone does not address, resulting in the best performance.

\begin{figure}[H]
    \centering
    \begin{minipage}[b]{0.14\textwidth}
        \centering
        \resizebox{\textwidth}{!}{
        \begin{tabular}{lc}
        \toprule
        \textbf{Method} & \makecell{Dynamic \\Attribution ($\uparrow$)}\\
        \cmidrule{1-2}
           LaVie  & 0.01242 \\
           + GPT & 0.01360 \\
        \cmidrule{1-2}
           VC2 & 0.00663\\
           + GPT &  0.00770\\
        \bottomrule
        \end{tabular}
        }
        \captionof{table}{Results for T2V-CompBench.}
        \label{tab4}
    \end{minipage}
    \hfill
    \begin{minipage}[b]{0.3\textwidth}
    \centering
    \includegraphics[width=\linewidth]{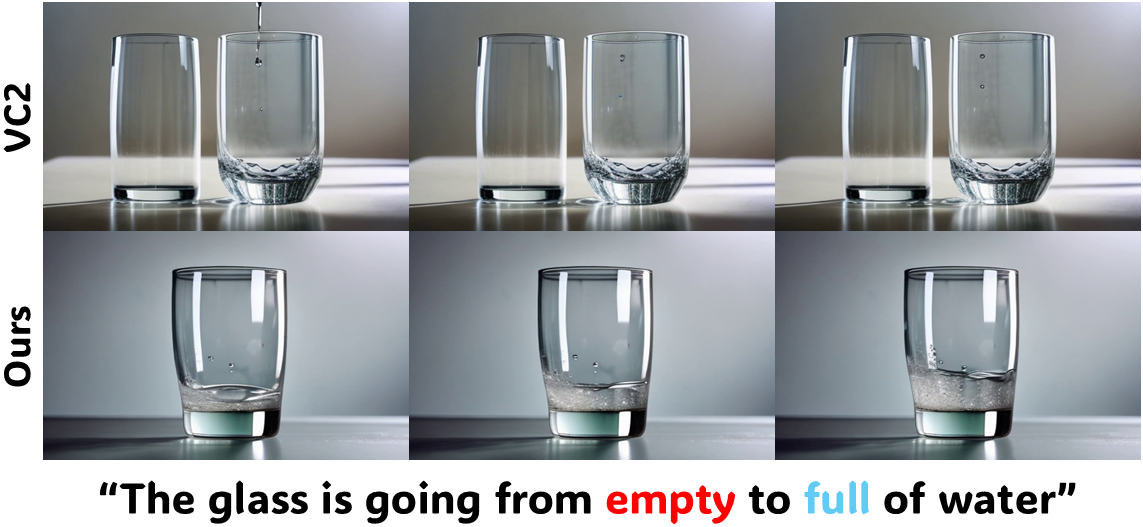}
    \caption{Example of T2V-CompBench.}
    \label{fig1}
    \end{minipage}
    \vspace{-0.4cm}
\end{figure}

As shown in Table~\ref{tab4}, leveraging LVLM improves performance in Dynamic Attribution Binding. Figure~\ref{fig1} illustrates an example video where the water gradually fills up over time in response to a given prompt when utilizing LVLM, whereas the baseline model fails to capture this progression.

\begin{figure}[H]
    \centering
    \begin{minipage}[b]{0.23\textwidth}
        \centering
        \resizebox{\textwidth}{!}{
        \begin{tabular}{lcc}
        \toprule
        \textbf{Method} &  \textbf{NFEs} & \textbf{Avg.}\\
        \cmidrule{1-3}
        Baseline & 100 & 0.5815 \\ 
        Best-of-N & 100 & 0.5802\\
        Ours & 100 & \textbf{0.5981} \\        
        \bottomrule
        \end{tabular}
        }
        \captionof{table}{Fixed NFE comparison on VBench.}
        \label{tab1}
    \end{minipage}
    \hfill
    \begin{minipage}[b]{0.22\textwidth}
    \centering
        \resizebox{1.0\textwidth}{!}{
        \begin{tabular}{lccc}
            \toprule
            Method  & CLIP ($\uparrow$)& IR ($\uparrow$) & GPT ($\uparrow$)\\
            \midrule
            \rowcolor{LightCyan} 
            VC2 & 30.39& -0.10 & 7.09\\
            +GPT & 30.90& \underline{0.23}& \underline{7.28}\\
            \midrule
            +CLIP & \textbf{30.96} & 0.14 &7.11\\
            ++GPT & \underline{30.95} & 0.20 &7.07\\
            \midrule
            +IR & 30.92 & 0.22 &\underline{7.28}\\
            ++GPT & \textbf{30.96}& \textbf{0.28} &\textbf{7.33}\\
            \bottomrule
        \end{tabular}
        }
        \captionof{table}{Reward robustness.}
        \label{tab:robustness}
    \end{minipage}
\end{figure}

\vspace{-0.2cm}
\subsection{Analysis}

\paragraph{Computational efficiency}
To evaluate the computational efficiency of our method, we conduct experiments under a fixed NFE budget of 100 using VideoCrafter2, as shown in Table \ref{tab1}. The Baseline uses a single 100-step inference path, while Best-of-N selects the highest LVLM reward from two 50-step paths. Our approach uses 50 steps, with six samples in the first 10 steps, while the remaining 40 steps follow the baseline procedure. Notably, simply selecting from multiple final outputs is ineffective, as it does not influence the denoising process. In contrast, our method actively guides generation throughout sampling, leading to improved text alignment and coherence that cannot be achieved through post-hoc selection.

\vspace*{-0.4cm}
\paragraph{Robustness of Rewards}
We verify that our method achieves robust performance without overfitting to any particular reward, avoiding reward hacking, a common issue in RL literature. Table \ref{tab:robustness} compares the rewards for the final video outputs generated by each method. Video guidance ensembled with LVLM generally achieves higher metrics, exhibiting a trend similar to the text alignment results in Table \ref{tab:results_t2v}. These findings indicate that the ensemble approach is not over-optimized for a particular reward, enhancing robustness across diverse evaluation criteria.
Additional ablation studies can be found in Appendix~\ref{app.C}.

\section{Conclusion}
\label{sec:Conclusion}
In this paper, we introduced Free$^2$Guide, a novel gradient-free framework to enhance text-video alignment in diffusion-based generative models without relying on reward gradients. By approximating the gradient of the reward function, Free$^2$Guide effectively integrates non-differentiable reward models, including powerful black-box LVLMs, to steer the video generation process towards better alignment. Our experiments demonstrate that Free$^2$Guide consistently improves alignment with text prompts and general video quality. By enabling ensembling with LVLM, our method benefits from synergistic effects, further enhancing performance.

\section{Acknowledgments}

This work was supported by the National Research Foundation of Korea under Grant
RS-2024-00336454, and by the Institute of Information \& Communications Technology Planning \& Evaluation (IITP) grant funded by the Korean government~(MSIT) (No. RS-2024-00457882, AI Research Hub Project; No. RS-2025-02304967, AI Star Fellowship~(KAIST))

{
    \small
    \bibliographystyle{ieeenat_fullname}
    \bibliography{main}

\begin{thebibliography}{52}
\providecommand{\natexlab}[1]{#1}
\providecommand{\url}[1]{\texttt{#1}}
\expandafter\ifx\csname urlstyle\endcsname\relax
  \providecommand{\doi}[1]{doi: #1}\else
  \providecommand{\doi}{doi: \begingroup \urlstyle{rm}\Url}\fi

\bibitem[Achiam et~al.(2023)Achiam, Adler, Agarwal, Ahmad, Akkaya, Aleman, Almeida, Altenschmidt, Altman, Anadkat, et~al.]{achiam2023gpt}
Josh Achiam, Steven Adler, Sandhini Agarwal, Lama Ahmad, Ilge Akkaya, Florencia~Leoni Aleman, Diogo Almeida, Janko Altenschmidt, Sam Altman, Shyamal Anadkat, et~al.
\newblock Gpt-4 technical report.
\newblock \emph{arXiv preprint arXiv:2303.08774}, 2023.

\bibitem[Black et~al.(2023)Black, Janner, Du, Kostrikov, and Levine]{black2023training}
Kevin Black, Michael Janner, Yilun Du, Ilya Kostrikov, and Sergey Levine.
\newblock Training diffusion models with reinforcement learning.
\newblock \emph{arXiv preprint arXiv:2305.13301}, 2023.

\bibitem[Chen et~al.(2023)Chen, Xia, He, Zhang, Cun, Yang, Xing, Liu, Chen, Wang, Weng, and Shan]{chen2023videocrafter1}
Haoxin Chen, Menghan Xia, Yingqing He, Yong Zhang, Xiaodong Cun, Shaoshu Yang, Jinbo Xing, Yaofang Liu, Qifeng Chen, Xintao Wang, Chao Weng, and Ying Shan.
\newblock Videocrafter1: Open diffusion models for high-quality video generation, 2023.

\bibitem[Chen et~al.(2024)Chen, Zhang, Cun, Xia, Wang, Weng, and Shan]{chen2024videocrafter2}
Haoxin Chen, Yong Zhang, Xiaodong Cun, Menghan Xia, Xintao Wang, Chao Weng, and Ying Shan.
\newblock Videocrafter2: Overcoming data limitations for high-quality video diffusion models, 2024.

\bibitem[Chung et~al.(2023)Chung, Kim, Mccann, Klasky, and Ye]{chung2023diffusion}
Hyungjin Chung, Jeongsol Kim, Michael~Thompson Mccann, Marc~Louis Klasky, and Jong~Chul Ye.
\newblock Diffusion posterior sampling for general noisy inverse problems.
\newblock In \emph{International Conference on Learning Representations}, 2023.

\bibitem[Clark et~al.(2023)Clark, Vicol, Swersky, and Fleet]{clark2023directly}
Kevin Clark, Paul Vicol, Kevin Swersky, and David~J Fleet.
\newblock Directly fine-tuning diffusion models on differentiable rewards.
\newblock \emph{arXiv preprint arXiv:2309.17400}, 2023.

\bibitem[Dhariwal and Nichol(2021)]{dhariwal2021diffusion}
Prafulla Dhariwal and Alexander~Quinn Nichol.
\newblock Diffusion models beat {GAN}s on image synthesis.
\newblock In \emph{Advances in Neural Information Processing Systems}, 2021.

\bibitem[Efron(2011)]{efron2011tweedie}
Bradley Efron.
\newblock Tweedie’s formula and selection bias.
\newblock \emph{Journal of the American Statistical Association}, 106\penalty0 (496):\penalty0 1602--1614, 2011.

\bibitem[Emerson(2013)]{emerson2013original}
Peter Emerson.
\newblock The original borda count and partial voting.
\newblock \emph{Social Choice and Welfare}, 40\penalty0 (2):\penalty0 353--358, 2013.

\bibitem[Fan et~al.(2024)Fan, Watkins, Du, Liu, Ryu, Boutilier, Abbeel, Ghavamzadeh, Lee, and Lee]{fan2024reinforcement}
Ying Fan, Olivia Watkins, Yuqing Du, Hao Liu, Moonkyung Ryu, Craig Boutilier, Pieter Abbeel, Mohammad Ghavamzadeh, Kangwook Lee, and Kimin Lee.
\newblock Reinforcement learning for fine-tuning text-to-image diffusion models.
\newblock \emph{Advances in Neural Information Processing Systems}, 36, 2024.

\bibitem[Feng et~al.(2024)Feng, Zhu, Fu, Jampani, Akula, He, Basu, Wang, and Wang]{feng2024layoutgpt}
Weixi Feng, Wanrong Zhu, Tsu-jui Fu, Varun Jampani, Arjun Akula, Xuehai He, Sugato Basu, Xin~Eric Wang, and William~Yang Wang.
\newblock Layoutgpt: Compositional visual planning and generation with large language models.
\newblock \emph{Advances in Neural Information Processing Systems}, 36, 2024.

\bibitem[Gokhale et~al.(2022)Gokhale, Palangi, Nushi, Vineet, Horvitz, Kamar, Baral, and Yang]{gokhale2022benchmarking}
Tejas Gokhale, Hamid Palangi, Besmira Nushi, Vibhav Vineet, Eric Horvitz, Ece Kamar, Chitta Baral, and Yezhou Yang.
\newblock Benchmarking spatial relationships in text-to-image generation.
\newblock \emph{arXiv preprint arXiv:2212.10015}, 2022.

\bibitem[He et~al.(2022)He, Yang, Zhang, Shan, and Chen]{he2022latent}
Yingqing He, Tianyu Yang, Yong Zhang, Ying Shan, and Qifeng Chen.
\newblock Latent video diffusion models for high-fidelity long video generation.
\newblock \emph{arXiv preprint arXiv:2211.13221}, 2022.

\bibitem[Ho and Salimans(2022)]{ho2022classifier}
Jonathan Ho and Tim Salimans.
\newblock Classifier-free diffusion guidance.
\newblock \emph{arXiv preprint arXiv:2207.12598}, 2022.

\bibitem[Ho et~al.(2020)Ho, Jain, and Abbeel]{ho2020denoising}
Jonathan Ho, Ajay Jain, and Pieter Abbeel.
\newblock Denoising diffusion probabilistic models.
\newblock \emph{Advances in Neural Information Processing Systems}, 33:\penalty0 6840--6851, 2020.

\bibitem[Ho et~al.(2022)Ho, Salimans, Gritsenko, Chan, Norouzi, and Fleet]{ho2022video}
Jonathan Ho, Tim Salimans, Alexey Gritsenko, William Chan, Mohammad Norouzi, and David~J Fleet.
\newblock Video diffusion models.
\newblock \emph{Advances in Neural Information Processing Systems}, 35:\penalty0 8633--8646, 2022.

\bibitem[Huang et~al.(2024{\natexlab{a}})Huang, Ghatare, Liu, Hu, Zhang, Sastry, Gururani, Oore, and Yue]{huang2024symbolic}
Yujia Huang, Adishree Ghatare, Yuanzhe Liu, Ziniu Hu, Qinsheng Zhang, Chandramouli~S Sastry, Siddharth Gururani, Sageev Oore, and Yisong Yue.
\newblock Symbolic music generation with non-differentiable rule guided diffusion.
\newblock \emph{arXiv preprint arXiv:2402.14285}, 2024{\natexlab{a}}.

\bibitem[Huang et~al.(2024{\natexlab{b}})Huang, He, Yu, Zhang, Si, Jiang, Zhang, Wu, Jin, Chanpaisit, Wang, Chen, Wang, Lin, Qiao, and Liu]{huang2023vbench}
Ziqi Huang, Yinan He, Jiashuo Yu, Fan Zhang, Chenyang Si, Yuming Jiang, Yuanhan Zhang, Tianxing Wu, Qingyang Jin, Nattapol Chanpaisit, Yaohui Wang, Xinyuan Chen, Limin Wang, Dahua Lin, Yu Qiao, and Ziwei Liu.
\newblock {VBench}: Comprehensive benchmark suite for video generative models.
\newblock In \emph{Proceedings of the IEEE/CVF Conference on Computer Vision and Pattern Recognition}, 2024{\natexlab{b}}.

\bibitem[Kaplan et~al.(2020)Kaplan, McCandlish, Henighan, Brown, Chess, Child, Gray, Radford, Wu, and Amodei]{kaplan2020scaling}
Jared Kaplan, Sam McCandlish, Tom Henighan, Tom~B Brown, Benjamin Chess, Rewon Child, Scott Gray, Alec Radford, Jeffrey Wu, and Dario Amodei.
\newblock Scaling laws for neural language models.
\newblock \emph{arXiv preprint arXiv:2001.08361}, 2020.

\bibitem[Kappen(2005)]{kappen2005path}
Hilbert~J Kappen.
\newblock Path integrals and symmetry breaking for optimal control theory.
\newblock \emph{Journal of statistical mechanics: theory and experiment}, 2005\penalty0 (11):\penalty0 P11011, 2005.

\bibitem[Karras et~al.(2022)Karras, Aittala, Aila, and Laine]{karras2022elucidating}
Tero Karras, Miika Aittala, Timo Aila, and Samuli Laine.
\newblock Elucidating the design space of diffusion-based generative models.
\newblock In \emph{Proc. NeurIPS}, 2022.

\bibitem[Kim et~al.(2024)Kim, Choi, Lee, and Rhee]{kim2024image}
Wonkyun Kim, Changin Choi, Wonseok Lee, and Wonjong Rhee.
\newblock An image grid can be worth a video: Zero-shot video question answering using a vlm.
\newblock \emph{IEEE Access}, 2024.

\bibitem[Kojima et~al.(2022)Kojima, Gu, Reid, Matsuo, and Iwasawa]{kojima2022large}
Takeshi Kojima, Shixiang~Shane Gu, Machel Reid, Yutaka Matsuo, and Yusuke Iwasawa.
\newblock Large language models are zero-shot reasoners.
\newblock \emph{Advances in neural information processing systems}, 35:\penalty0 22199--22213, 2022.

\bibitem[Li et~al.(2024)]{li2024mvbench}
K Li et~al.
\newblock Mvbench: A comprehensive multi-modal video understanding benchmark.
\newblock In \emph{CVPR}, 2024.

\bibitem[Lian et~al.(2023)Lian, Shi, Yala, Darrell, and Li]{lian2023llm}
Long Lian, Baifeng Shi, Adam Yala, Trevor Darrell, and Boyi Li.
\newblock Llm-grounded video diffusion models.
\newblock \emph{arXiv preprint arXiv:2309.17444}, 2023.

\bibitem[Liu et~al.(2020)Liu, Chen, Kailkhura, Zhang, Hero~III, and Varshney]{liu2020primer}
Sijia Liu, Pin-Yu Chen, Bhavya Kailkhura, Gaoyuan Zhang, Alfred~O Hero~III, and Pramod~K Varshney.
\newblock A primer on zeroth-order optimization in signal processing and machine learning: Principals, recent advances, and applications.
\newblock \emph{IEEE Signal Processing Magazine}, 37\penalty0 (5):\penalty0 43--54, 2020.

\bibitem[Luo et~al.(2023)Luo, Tan, Huang, Li, and Zhao]{luo2023latent}
Simian Luo, Yiqin Tan, Longbo Huang, Jian Li, and Hang Zhao.
\newblock Latent consistency models: Synthesizing high-resolution images with few-step inference.
\newblock \emph{arXiv preprint arXiv:2310.04378}, 2023.

\bibitem[Nesterov and Spokoiny(2017)]{nesterov2017random}
Yurii Nesterov and Vladimir Spokoiny.
\newblock Random gradient-free minimization of convex functions.
\newblock \emph{Foundations of Computational Mathematics}, 17\penalty0 (2):\penalty0 527--566, 2017.

\bibitem[Nie et~al.(2022)Nie, Guo, Huang, Xiao, Vahdat, and Anandkumar]{nie2022diffusion}
Weili Nie, Brandon Guo, Yujia Huang, Chaowei Xiao, Arash Vahdat, and Anima Anandkumar.
\newblock Diffusion models for adversarial purification.
\newblock \emph{arXiv preprint arXiv:2205.07460}, 2022.

\bibitem[Prabhudesai et~al.(2023)Prabhudesai, Goyal, Pathak, and Fragkiadaki]{prabhudesai2023aligning}
Mihir Prabhudesai, Anirudh Goyal, Deepak Pathak, and Katerina Fragkiadaki.
\newblock Aligning text-to-image diffusion models with reward backpropagation.
\newblock \emph{arXiv preprint arXiv:2310.03739}, 2023.

\bibitem[Prabhudesai et~al.(2024)Prabhudesai, Mendonca, Qin, Fragkiadaki, and Pathak]{prabhudesai2024video}
Mihir Prabhudesai, Russell Mendonca, Zheyang Qin, Katerina Fragkiadaki, and Deepak Pathak.
\newblock Video diffusion alignment via reward gradients.
\newblock \emph{arXiv preprint arXiv:2407.08737}, 2024.

\bibitem[Radford et~al.(2021)Radford, Kim, Hallacy, Ramesh, Goh, Agarwal, Sastry, Askell, Mishkin, Clark, et~al.]{radford2021learning}
Alec Radford, Jong~Wook Kim, Chris Hallacy, Aditya Ramesh, Gabriel Goh, Sandhini Agarwal, Girish Sastry, Amanda Askell, Pamela Mishkin, Jack Clark, et~al.
\newblock Learning transferable visual models from natural language supervision.
\newblock In \emph{International conference on machine learning}, pages 8748--8763. PMLR, 2021.

\bibitem[Rombach et~al.(2022)Rombach, Blattmann, Lorenz, Esser, and Ommer]{rombach2022high}
Robin Rombach, Andreas Blattmann, Dominik Lorenz, Patrick Esser, and Bj{\"o}rn Ommer.
\newblock High-resolution image synthesis with latent diffusion models.
\newblock In \emph{Proceedings of the IEEE/CVF Conference on Computer Vision and Pattern Recognition}, pages 10684--10695, 2022.

\bibitem[Sohl-Dickstein et~al.(2015)Sohl-Dickstein, Weiss, Maheswaranathan, and Ganguli]{sohl2015deep}
Jascha Sohl-Dickstein, Eric Weiss, Niru Maheswaranathan, and Surya Ganguli.
\newblock Deep unsupervised learning using nonequilibrium thermodynamics.
\newblock In \emph{International Conference on Machine Learning}, pages 2256--2265. PMLR, 2015.

\bibitem[Song et~al.(2021{\natexlab{a}})Song, Meng, and Ermon]{song2020denoising}
Jiaming Song, Chenlin Meng, and Stefano Ermon.
\newblock Denoising diffusion implicit models.
\newblock In \emph{9th International Conference on Learning Representations, {ICLR}}, 2021{\natexlab{a}}.

\bibitem[Song et~al.(2021{\natexlab{b}})Song, Sohl-Dickstein, Kingma, Kumar, Ermon, and Poole]{song2021scorebased}
Yang Song, Jascha Sohl-Dickstein, Diederik~P Kingma, Abhishek Kumar, Stefano Ermon, and Ben Poole.
\newblock Score-based generative modeling through stochastic differential equations.
\newblock In \emph{International Conference on Learning Representations}, 2021{\natexlab{b}}.

\bibitem[Subramanian et~al.(2022)Subramanian, Merrill, Darrell, Gardner, Singh, and Rohrbach]{subramanian2022reclip}
Sanjay Subramanian, William Merrill, Trevor Darrell, Matt Gardner, Sameer Singh, and Anna Rohrbach.
\newblock Reclip: A strong zero-shot baseline for referring expression comprehension.
\newblock \emph{arXiv preprint arXiv:2204.05991}, 2022.

\bibitem[Sun et~al.(2024)]{sun2024t2v}
K Sun et~al.
\newblock T2v-compbench: A comprehensive benchmark for compositional text-to-video generation.
\newblock \emph{arXiv}, 2024.

\bibitem[Theodorou et~al.(2010)Theodorou, Buchli, and Schaal]{theodorou2010generalized}
Evangelos Theodorou, Jonas Buchli, and Stefan Schaal.
\newblock A generalized path integral control approach to reinforcement learning.
\newblock \emph{The Journal of Machine Learning Research}, 11:\penalty0 3137--3181, 2010.

\bibitem[Uehara et~al.(2024{\natexlab{a}})Uehara, Zhao, Biancalani, and Levine]{uehara2024understanding}
Masatoshi Uehara, Yulai Zhao, Tommaso Biancalani, and Sergey Levine.
\newblock Understanding reinforcement learning-based fine-tuning of diffusion models: A tutorial and review.
\newblock \emph{arXiv preprint arXiv:2407.13734}, 2024{\natexlab{a}}.

\bibitem[Uehara et~al.(2024{\natexlab{b}})Uehara, Zhao, Black, Hajiramezanali, Scalia, Diamant, Tseng, Biancalani, and Levine]{uehara2024fine}
Masatoshi Uehara, Yulai Zhao, Kevin Black, Ehsan Hajiramezanali, Gabriele Scalia, Nathaniel~Lee Diamant, Alex~M Tseng, Tommaso Biancalani, and Sergey Levine.
\newblock Fine-tuning of continuous-time diffusion models as entropy-regularized control.
\newblock \emph{arXiv preprint arXiv:2402.15194}, 2024{\natexlab{b}}.

\bibitem[Wallace et~al.(2023)Wallace, Gokul, Ermon, and Naik]{wallace2023end}
Bram Wallace, Akash Gokul, Stefano Ermon, and Nikhil Naik.
\newblock End-to-end diffusion latent optimization improves classifier guidance.
\newblock In \emph{Proceedings of the IEEE/CVF International Conference on Computer Vision}, pages 7280--7290, 2023.

\bibitem[Wang et~al.(2023{\natexlab{a}})Wang, Chen, Ma, Zhou, Huang, Wang, Yang, He, Yu, Yang, et~al.]{wang2023lavie}
Yaohui Wang, Xinyuan Chen, Xin Ma, Shangchen Zhou, Ziqi Huang, Yi Wang, Ceyuan Yang, Yinan He, Jiashuo Yu, Peiqing Yang, et~al.
\newblock Lavie: High-quality video generation with cascaded latent diffusion models.
\newblock \emph{arXiv preprint arXiv:2309.15103}, 2023{\natexlab{a}}.

\bibitem[Wang et~al.(2023{\natexlab{b}})Wang, He, Li, Li, Yu, Ma, Li, Chen, Chen, Wang, et~al.]{wang2023internvid}
Yi Wang, Yinan He, Yizhuo Li, Kunchang Li, Jiashuo Yu, Xin Ma, Xinhao Li, Guo Chen, Xinyuan Chen, Yaohui Wang, et~al.
\newblock Internvid: A large-scale video-text dataset for multimodal understanding and generation.
\newblock In \emph{The Twelfth International Conference on Learning Representations}, 2023{\natexlab{b}}.

\bibitem[Williams and Rogers(1979)]{williams1979diffusions}
David Williams and L~Chris~G Rogers.
\newblock \emph{Diffusions, Markov processes, and martingales}.
\newblock John Wiley \& Sons, 1979.

\bibitem[Wu et~al.(2024)Wu, Lian, Gonzalez, Li, and Darrell]{wu2024self}
Tsung-Han Wu, Long Lian, Joseph~E Gonzalez, Boyi Li, and Trevor Darrell.
\newblock Self-correcting llm-controlled diffusion models.
\newblock In \emph{Proceedings of the IEEE/CVF Conference on Computer Vision and Pattern Recognition}, pages 6327--6336, 2024.

\bibitem[Wu et~al.(2023)Wu, Hao, Sun, Chen, Zhu, Zhao, and Li]{wu2023human}
Xiaoshi Wu, Yiming Hao, Keqiang Sun, Yixiong Chen, Feng Zhu, Rui Zhao, and Hongsheng Li.
\newblock Human preference score v2: A solid benchmark for evaluating human preferences of text-to-image synthesis.
\newblock \emph{arXiv preprint arXiv:2306.09341}, 2023.

\bibitem[Xu et~al.(2024)Xu, Liu, Wu, Tong, Li, Ding, Tang, and Dong]{xu2024imagereward}
Jiazheng Xu, Xiao Liu, Yuchen Wu, Yuxuan Tong, Qinkai Li, Ming Ding, Jie Tang, and Yuxiao Dong.
\newblock Imagereward: Learning and evaluating human preferences for text-to-image generation.
\newblock \emph{Advances in Neural Information Processing Systems}, 36, 2024.

\bibitem[Yang et~al.(2024)Yang, Yu, Meng, Xu, Ermon, and Bin]{yang2024mastering}
Ling Yang, Zhaochen Yu, Chenlin Meng, Minkai Xu, Stefano Ermon, and CUI Bin.
\newblock Mastering text-to-image diffusion: Recaptioning, planning, and generating with multimodal llms.
\newblock In \emph{Forty-first International Conference on Machine Learning}, 2024.

\bibitem[Yeh et~al.(2024)Yeh, Lee, and Chen]{yeh2024training}
Po-Hung Yeh, Kuang-Huei Lee, and Jun-Cheng Chen.
\newblock Training-free diffusion model alignment with sampling demons.
\newblock \emph{arXiv preprint arXiv:2410.05760}, 2024.

\bibitem[Zheng et~al.(2024)Zheng, Chu, Wang, Kovachki, Baptista, and Yue]{zheng2024ensemble}
Hongkai Zheng, Wenda Chu, Austin Wang, Nikola Kovachki, Ricardo Baptista, and Yisong Yue.
\newblock Ensemble kalman diffusion guidance: A derivative-free method for inverse problems.
\newblock \emph{arXiv preprint arXiv:2409.20175}, 2024.

\bibitem[Zhong et~al.(2023)Zhong, Huang, Wen, Qin, and Lin]{zhong2023adapter}
Shanshan Zhong, Zhongzhan Huang, Weushao Wen, Jinghui Qin, and Liang Lin.
\newblock Sur-adapter: Enhancing text-to-image pre-trained diffusion models with large language models.
\newblock In \emph{Proceedings of the 31st ACM International Conference on Multimedia}, pages 567--578, 2023.

\end{thebibliography}
}


\clearpage
\onecolumn 
\appendix
\setcounter{page}{1}    

\section{Implementation Details}
\label{sec:implementation}

\subsection{Model Checkpoints}

We use the pre-trained T2V diffusion model LaVie and VideoCrafter2, available at \url{https://github.com/Vchitect/LaVie} and \url{https://github.com/AILab-CVC/VideoCrafter}, respectively. For LaVie, the Stable Diffusion v1.4 model is employed to encode and decode latent. We also utilize CLIP from \url{https://huggingface.co/openai/clip-vit-base-patch32} and the ImageReward model from \url{https://github.com/THUDM/ImageReward}.

\subsection{Evaluation Details}

During the video guidance process, we extract key frames from the video—specifically, the first, sixth, eleventh, and sixteenth frames—and assess the reward. When using an LVLM as the reward model, we concatenate the key frames using the following scripts:

\definecolor{codegreen}{rgb}{0,0.6,0}
\definecolor{codegray}{rgb}{0.5,0.5,0.5}
\definecolor{codepurple}{rgb}{0.58,0,0.82}
\definecolor{backcolour}{rgb}{0.95,0.95,0.92}

\lstdefinestyle{mystyle}{
    backgroundcolor=\color{backcolour},   
    commentstyle=\color{codegreen},
    keywordstyle=\color{magenta},
    numberstyle=\tiny\color{codegray},
    stringstyle=\color{codepurple},
    basicstyle=\ttfamily\footnotesize,
    breakatwhitespace=false,         
    breaklines=true,                 
    captionpos=b,                    
    keepspaces=true,                 
    numbers=left,                    
    numbersep=5pt,                  
    showspaces=false,                
    showstringspaces=false,
    showtabs=false,        
    frame=single,
    tabsize=2
}

\lstset{style=mystyle}

\begin{lstlisting}[language=Python, caption={Pseudo-code for stitching key frames at once.}]
fig, axes = plt.subplots(2, 2, figsize=(12, 8))
key_frames = [0, 5, 10, 15]

for idx, frame in enumerate(key_frames):
    ax = axes[idx // 2, idx % 2]
    ax.imshow(video[0, :, frame, :, :].permute(1, 2, 0).cpu().numpy())
    ax.axis('off')
    ax.set_title(f'Frame {frame + 1}')

# Adjust the layout and show the plot
plt.tight_layout()
plt.savefig(f'frame_{i}_{j}.png')
    
\end{lstlisting}

Next, we provide a system instruction that allows the LVLM to understand the sequence order and explicitly describes the task it should perform.

\lstset{
    basicstyle=\small,
    breaklines=true,
    columns=flexible,
    frame=single,
    firstnumber=1
}
\begin{lstlisting}[language={}, label={lst:script1}, caption={System instruction for GPT-4o}]
You are a useful helper that responds to video quality assessments.
The given image is a grid of four key frames of a video: the top left is the first frame, the top right is the second 
frame, the bottom left is the third frame, and finally the bottom right is the fourth frame. 
Answer the reason first and the final answer later. Start the reason first with `Reasoning: ' in front of the reason part 
and review your reasoning logically.
After reviewing your reasoning, give the final answer with `Answer: '. 
You should check all frame and comparing them, and ensure your reasoning leads to a sound final answer.
Your final `answer' should one score only and the score must be from 1 to 9 without decimals.
Let's think step by step.
\end{lstlisting}

For a given video, we input the user prompt to the LVLM as follows:

\begin{lstlisting}[language={}, label={lst:script2}, caption={User prompt for GPT-4o}, escapeinside=``]
For a given image as keyframes of video, Rate the following questions:
Considering all four images, does the prompt, `\textbf{\texttt{prompt}}`, describe the video well enough?
Review your reasoning thoroughly and then respond with your final decision prefixed by `Answer: '.
\end{lstlisting}

where $\textbf{\texttt{prompt}}$ is the given text prompt (\eg ``a bird and a cat'')

\section{Limitation} 
Sampling in our approach requires additional processing time to approximate the gradient. While our approach extends sampling time compared to baseline, it uniquely enables guidance with non-differentiable reward models such as LVLM APIs. Additionally, the effectiveness of our framework is influenced by the accuracy of the reward function, which opens avenues for further improvements as reward models continue to advance.

\section{Additional Ablation Study}
\label{app.C}

\begin{table}[H]
\centering
\begin{minipage}[h]{0.65\textwidth}
    \paragraph{Number of Samples} We analyze the effect of the sampling quantity on text alignment performance, evaluating the average text alignment score using the LaVie model with a CLIP reward model. As shown in Table \ref{tab:abl1}, we find an optimal sampling size at $n = 5$. Increasing the number of samples increases the likelihood of selecting a denoised video that aligns with the desired control. However, excessive sampling introduces a risk: errors predicted by Tweedie's formula in initial sampling steps may result in irreversible changes, affecting video quality negatively.
\end{minipage} \hfill
\begin{minipage}[h]{0.3\textwidth}
\centering
\vspace{-0.6cm}
        \begin{tabular}{ccc}
            \toprule
            $n$  & Avg. \\
            \midrule
            1 & 0.3722 \\
            3 & 0.3749 \\
            5 & \textbf{0.3780} \\
            10 & 0.3705 \\
            \bottomrule
        \end{tabular}
        \caption{Quantitative results on text alignment by sample size.}
        \label{tab:abl1}    
        \vspace{-0.5cm}
\end{minipage}
\end{table}

\vspace{-0.4cm}

\begin{table}[H]
\centering
\begin{minipage}[h]{0.65\textwidth}
    \paragraph{Guidance Range} We also evaluate the effect of the guidance range with the same baseline. Table \ref{tab:abl2} reveals that applying guidance in the early stages is more effective than in later stages, as these initial steps establish the overall spatial structure of the video. However, extending the guidance range too far allows errors in the approximated optimal control to accumulate, ultimately degrading the quality of the final output video.
\end{minipage} \hfill
\begin{minipage}[h]{0.3\textwidth}
\centering
        \begin{tabular}{lc}
            \toprule
            Guidance Step & Avg. \\
            \midrule
            None & 0.3722 \\
            $t \in [T,T-5]$ & \textbf{0.3780} \\
            $t \in [T-5,T-10]$ & 0.3769 \\
            $t \in [T,T-10]$ & 0.3635 \\
            \bottomrule
        \end{tabular}
        \caption{Quantitative results on text alignment by range of guidance step.}
        \label{tab:abl2}
        \vspace{-0.5cm}
\end{minipage}
\end{table}

\begin{table}[H]
\centering
\begin{minipage}[h]{0.65\textwidth}
    \paragraph{Assessment policy using LVLM} We evaluate the impact of the assessment protocol in LVLM by analyzing the average scores generated with the VideoCrafter2 model. Specifically, we modify the system prompt to instruct LVLM to answer only with `yes' or `no' when assessing text-video alignment. The alignment score is then derived by calculating the percentage of the top 5 logits that correspond to `yes'. Table \ref{tab:abl3} reveals that scoring alignment on a scale from 1 to 9 achieves better performance in terms of text alignment. This is likely because a broader scale allows for more nuanced distinctions in fidelity, enabling LVLM to capture subtle differences in text-video alignment more effectively.
\end{minipage} \hfill
\begin{minipage}[h]{0.3\textwidth}
    \centering
    \resizebox{1.0\linewidth}{!}{
    \begin{tabular}{lccc}
    \toprule
        Method & Text Alignment & General Quailty & Avg. \\
    \midrule
    \rowcolor{LightCyan} 
        VC2 & 0.4129& 0.7617& 0.5873 \\
        +GPT\textsubscript{0/1} & 0.4358& \textbf{0.7550}&0.5954 \\
        +GPT\textsubscript{1-9} & \textbf{0.4425}& 0.7537& \textbf{0.5981} \\
    \bottomrule
    \end{tabular}
    }
    \caption{Average results by assessment policy using LVLM.}
    \label{tab:abl3}
        \vspace{-0.5cm}
\end{minipage}
\end{table}

\section{Additional Analysis}

\begin{center}
\centering
\small
\resizebox{\linewidth}{!}{%
\begin{tabularx}{1.2\linewidth}{l>{\centering\arraybackslash}m{2cm} >{\centering\arraybackslash}m{2cm} >{\centering\arraybackslash}m{2cm} >{\centering\arraybackslash}m{2cm} >{\centering\arraybackslash}m{2cm} >{\centering\arraybackslash}m{2cm} >{\centering\arraybackslash}m{2cm}}
\toprule
 \multicolumn{1}{c}{\scalebox{1.2}{\textbf{Method}}} & \makecell{Appearance\\Style} & \makecell{Temporal\\Style} &   \makecell{Human\\Action} &   \makecell{Multiple\\Objects} &   \makecell{Spatial\\Relationship} &   \makecell{Overall\\Consistency} &  \multicolumn{1}{c}{\scalebox{1.2}{\textbf{Avg.}}}\\
\midrule
{\scalebox{1.0}{LaVie}} & 0.2312& \underline{0.2502}  & \underline{0.9300}  & 0.2027  & \underline{0.3496} & 0.2694 &0.3722  \\
{\scalebox{1.0}{+ GPT4o}} & \underline{0.2366}  &\textbf{0.2508}	 & \underline{0.9300}  & \textbf{0.2546}  &\textbf{0.3531}	 &\underline{0.2709}   & \textbf{0.3827}  \\
{\scalebox{1.0}{+ Qwen2.5-VL 3B Image}} &\textbf{0.2388}  &0.2447  &\textbf{0.9700}  &\underline{0.2477}  &0.3238  &0.2647   & \underline{0.3816}  \\
{\scalebox{1.0}{+ Qwen2.5-VL 3B Video}} & 0.2325& 0.2464& \textbf{0.9700}& 0.2431& 0.3101& \textbf{0.2738}& 0.3793 \\
\midrule
{\scalebox{1.0}{LTX-Video-2B}} & 0.2189& 0.1784& \textbf{0.5303}& 0.1994& 0.3436& 0.1916& 0.2770 \\
{\scalebox{1.0}{+ GPT4o}} & \textbf{0.2202}& \textbf{0.1813}& 0.5051& \textbf{0.2335}& \textbf{0.4177}& \textbf{0.1947}& \textbf{0.2921} \\
\bottomrule
\end{tabularx}
}
\captionof{table}{Baseline comparison with open-source Image and Video LVLM and longer video generation model.}
\label{tab:re_results_t2v}
\end{center}

\begin{figure}[h]
\centering

\begin{minipage}{0.45\linewidth}
\centering
\small
\begin{tabular}{lcc}
\toprule
\textbf{Aspects} & \textbf{Baselines} & \textbf{Ours}\\
\cmidrule{1-3}
Overall Quality & 2.61 & 3.19  \\ 
Temporal Quality & 2.65& 3.21\\
Text Alignment & 2.60& 3.94 \\        
\bottomrule
\end{tabular}
\captionof{table}{User study.}
\label{tab:re_userstudy}
\end{minipage}
\hfill
\begin{minipage}{0.45\linewidth}
\centering
\small
\begin{tabular}{lcc}
\toprule
\textbf{Method} & GPU Memory & Computing Time \\
\midrule
Lavie & 4.4 GiB & 22.7 s/video \\
\textbf{+Ours} & 7.5 GiB & 154.5 s/video \\
\bottomrule
\end{tabular}
\captionof{table}{Computation.}
\label{tab:re_overhead}
\end{minipage}

\end{figure}

\paragraph{Open-source LVLM.} We leverage an open-source LVLM (QWen2.5-VL 3B) using both stitched image input and direct video input. As shown in Table \ref{tab:re_results_t2v}, our framework consistently improves T2V alignment. Interestingly, image input demonstrated stronger performance than direct video input for this specific LVLM. We hypothesize this might be due to our frame stitching method effectively highlighting key temporal information for the LVLM.

\paragraph{Long Video Generation Model.} To address concerns about generalization to longer videos, we applied Free$^2$Guide to a long video generation model (LTX-video 2B), generating 15-second videos. As presented in Table \ref{tab:re_results_t2v}, we measure VBench2-beta-long metrics and our framework significantly improves performance over the baseline (which used stochastic sampling for fair comparison), demonstrating its effectiveness in longer videos.

\paragraph{User Study.} We conducted a user study with 50 participants on Prolific, comparing videos from our method against the baseline (LaVie and VideoCrafter2). Participants rated videos on a 1-5 scale for overall quality, temporal quality, and text alignment. Our method was consistently preferred across all aspects, as shown in Table \ref{tab:re_userstudy}.

\subsection{Video Reward Guidance}
\label{app.D}
While using a video-based reward model to guide videos is a more natural approach, we claim that video reward models fail to capture the representation needed for guidance because the dataset of video-text pairs is relatively limited compared to images. To support this, we compare the results of using a video-based reward model for guidance with a video-based reward model for text alignment. We adopt ViCLIP~\cite{wang2023internvid}, a pre-trained video-text representation learning model available at \url{https://huggingface.co/OpenGVLab/ViCLIP}, as the video reward model. Using LaVie as the baseline, we compute the reward based on eight video frames, measuring the similarity between the video and text embeddings. 

Table \ref{tab:viclip} shows that the video-based reward model does not significantly outperform the image-based reward model. However, it specifically enhances the Overall Consistency and Dynamic Degree metrics. It is worth noting that the Overall Consistency metric is evaluated using ViCLIP itself, which could introduce a bias favoring the video reward model. In addition, we observe that ViCLIP struggles with spatial information processing compared to CLIP, leading to lower performance on the Multiple Objects and Spatial Relationship metrics. These results highlight the challenges of video reward models to fully capture the relationship between video and text due to the lack of training datasets.

\begin{table*}[!t]
\centering
\renewcommand{\arraystretch}{1.1}
\resizebox{1.0\textwidth}{!}{
\begin{tabular}{lccccccc}
\toprule
 & \multicolumn{2}{c}{Style} & \multicolumn{3}{c}{Semantics} & \multicolumn{1}{c}{Condition Consistency} & \multirow{2}{*}{Avg.}\\
\cmidrule(lr){2-3} \cmidrule(lr){4-6} \cmidrule(lr){7-7}
 \multicolumn{1}{c}{\scalebox{1.2}{\textbf{Method}}} & {Appearance Style} & {Temporal Style} &   {Human Action} &   {Multiple Objects} &   {Spatial Relationship} &   {Overall Consistency} & \\
\midrule
\rowcolor{LightCyan} 
   {\scalebox{1.2}{LaVie \cite{wang2023lavie}}} & 0.2312& 0.2502  & 0.9300  & 0.2027  & 0.3496 & 0.2694   & 0.3722 \\
   {\scalebox{1.2}{+ CLIP}} & 0.2370 \red{\scalebox{0.8}{(+2.5\%)}}& 0.2490 \blue{\scalebox{0.8}{(-0.5\%)}} & 0.9400 \red{\scalebox{0.8}{(+1.1\%)}}& 0.2607 \red{\scalebox{0.8}{(+28.6\%)}}& 0.3074 \blue{\scalebox{0.8}{(-12.1\%)}}& 0.2738 \red{\scalebox{0.8}{(+1.6\%)}}& 0.3780\\   
  {\scalebox{1.2}{+ ViCLIP}} & 0.2348 \red{\scalebox{0.8}{(+1.6\%)}}& 0.2485 \blue{\scalebox{0.8}{(-0.7\%)}} & 0.9600 \red{\scalebox{0.8}{(+3.2\%)}}& 0.2149 \red{\scalebox{0.8}{(+6.0\%)}}& 0.2872 \blue{\scalebox{0.8}{(-17.9\%)}}& 0.2752 \red{\scalebox{0.8}{(+2.1\%)}}& 0.3701\\   
{\scalebox{1.2}{+ GPT}} & 0.2366 \red{\scalebox{0.8}{(+2.3\%)}}&0.2508	\red{\scalebox{0.8}{(+0.2\%)}}& 0.9300 \scalebox{0.8}{(-0.0\%)}& 0.2546 \red{\scalebox{0.8}{(+25.6\%)}}&0.3531	\red{\scalebox{0.8}{(+1.0\%)}}&0.2709 \red{\scalebox{0.8}{(+0.6\%)}}&0.3827 \\
\midrule

\toprule
 & \multicolumn{2}{c}{Temporal Consistency} & \multicolumn{2}{c}{Dynamics} & \multicolumn{2}{c}{Frame-wise Quality} & \multirow{2}{*}{Avg.} \\
\cmidrule(lr){2-3} \cmidrule(lr){4-5} \cmidrule(lr){6-7}
 \multicolumn{1}{c}{\scalebox{1.2}{\textbf{Method}}} & {Subject Consistency} & {Background Consistency} &   {Motion Smoothness} &   {Dynamic Degree} &   {Aesthetic Quality} &   {Imaging Quality} & \\
\midrule
\rowcolor{LightCyan} 
   {\scalebox{1.2}{LaVie \cite{wang2023lavie}}} & 0.9450& 0.9689  & 0.9718  & 0.4799  & 0.5687 & 0.6611 &0.7659   \\
  {\scalebox{1.2}{+ CLIP}} & 0.9495 \red{\scalebox{0.8}{(+0.5\%)}}& 0.9712 \red{\scalebox{0.8}{(+0.2\%)}} & 0.9735 \red{\scalebox{0.8}{(+0.2\%)}}& 0.4560 \blue{\scalebox{0.8}{(-5.0\%)}}& 0.5727 \red{\scalebox{0.8}{(0.7\%)}}& 0.6637 \red{\scalebox{0.8}{(+0.4\%)}} & 0.7644\\
   {\scalebox{1.2}{+ ViCLIP}} & 0.9443 \blue{\scalebox{0.8}{(-0.1\%)}}& 0.9694 \red{\scalebox{0.8}{(+0.0\%)}} & 0.9741 \red{\scalebox{0.8}{(+0.2\%)}}& 0.4707 \blue{\scalebox{0.8}{(-1.9\%)}}& 0.5746 \red{\scalebox{0.8}{(1.0\%)}}& 0.6487 \blue{\scalebox{0.8}{(-1.9\%)}} & 0.7636\\
 {\scalebox{1.2}{+ GPT}} & 0.9470 \red{\scalebox{0.8}{(+0.2\%)}}&0.9693	\red{\scalebox{0.8}{(+0.0\%)}}& 0.9742 \red{\scalebox{0.8}{(+0.2\%)}}& 0.4725 \blue{\scalebox{0.8}{(-1.5\%)}}&0.5726	\red{\scalebox{0.8}{(+0.7\%)}}&0.6615 \red{\scalebox{0.8}{(+0.1\%)}} & 0.7662  \\
\bottomrule

\end{tabular}
}
\caption{
Comparison with video-based reward model. Higher numbers indicate better video quality. The numbers in parentheses denote the performance difference from the baselines.
}
\label{tab:viclip}
\end{table*}

\subsection{Video Inverse Problems}

\begin{figure}[!b]
    \centering
    \includegraphics[width=\linewidth]{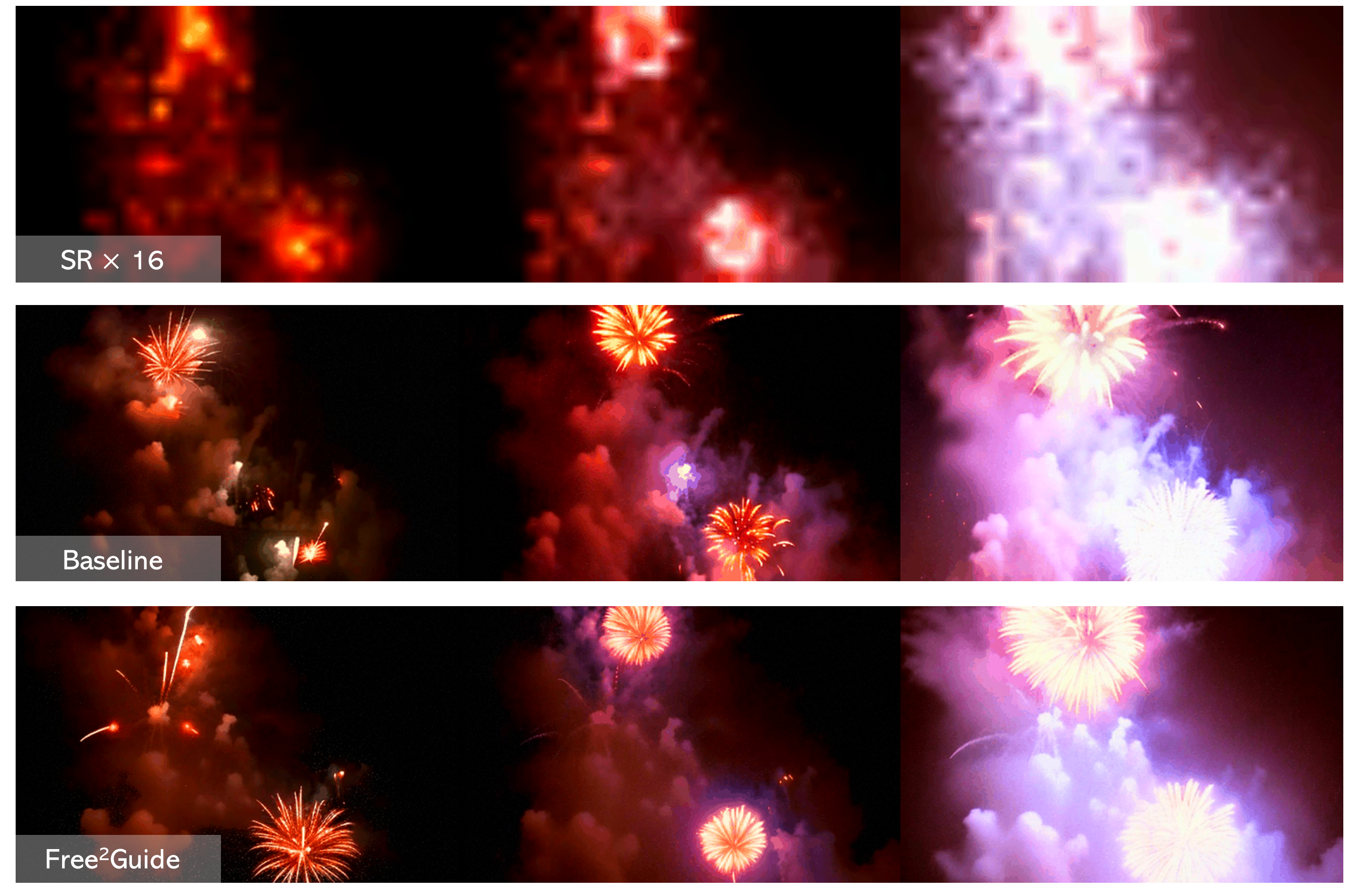}
    \caption{The result of applying our method to the inverse problem. Baseline represents that no guidance is applied during sampling.}
    \label{fig:inv}
\end{figure}

Our framework can readily extend to inverse problems in the video domain, building on approaches from previous work~\cite{zheng2024ensemble, huang2024symbolic}. In Figure~\ref{fig:inv}, we show a video reconstructed by our method using $\times$16 average pooling on spatial resolution. For the reward function, we use the $L_2$ distance between the corrupted denoised video and the corrupted video, applying a sampling size of 10 at each step with DDIM over 500 steps, using VideoCrafter2. Our results demonstrate that, compared to unguided sampling, our method generates realistic videos that remain faithful to the input. We leave further extension to video inverse problems as future work.

\newpage

\section{Additional Visual Results}

\begin{figure*}[!b]
    \centering
    \includegraphics[width=\linewidth]{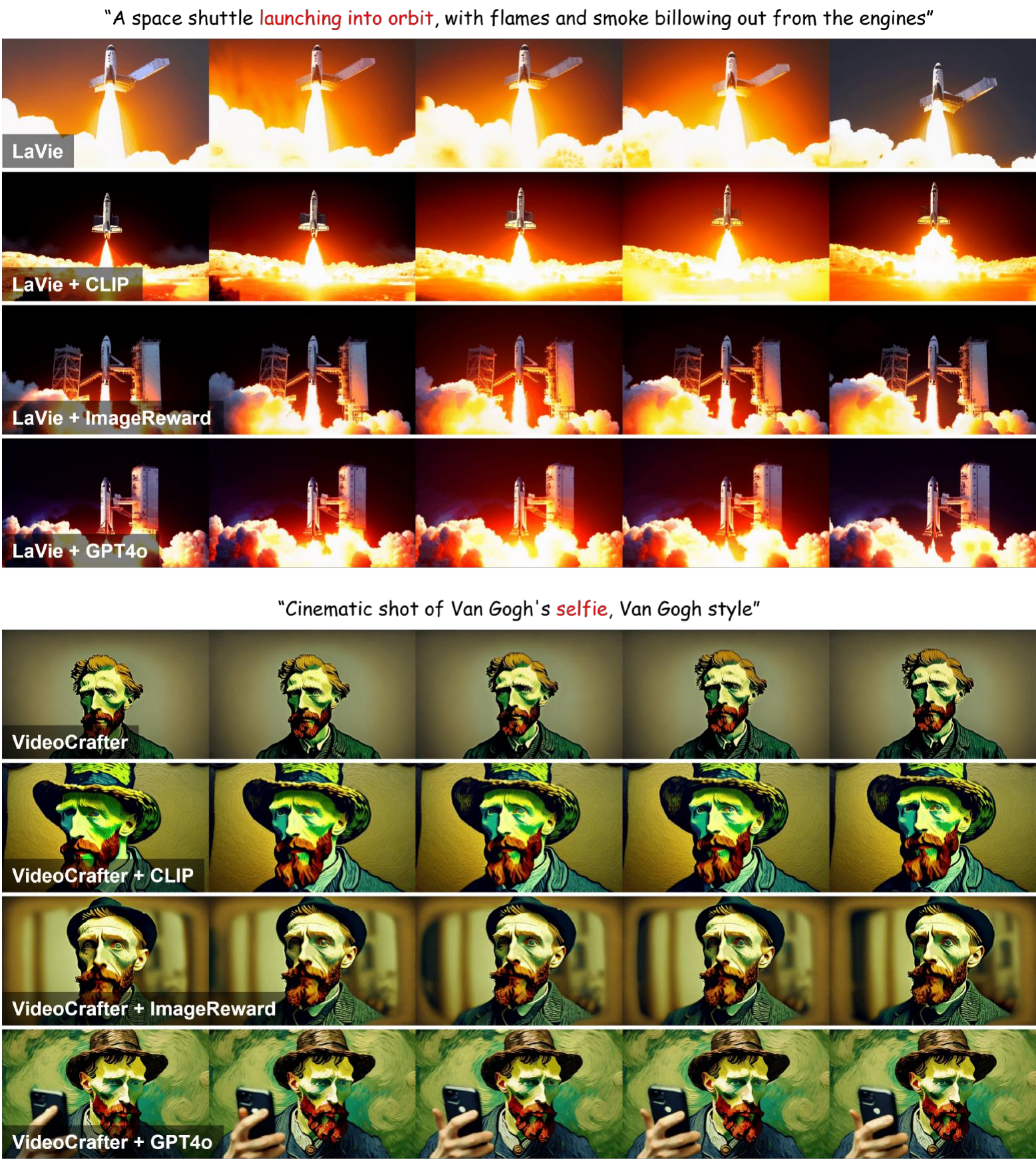}
    \vspace{-0.4cm}
    \caption{More qualitative comparison of different reward models. The red text highlights the difference between the models.
    }
    \label{fig:results_more1}
    \vspace{-0.2cm}
\end{figure*}

\begin{figure*}[!t]
    \centering
    \includegraphics[width=\linewidth]{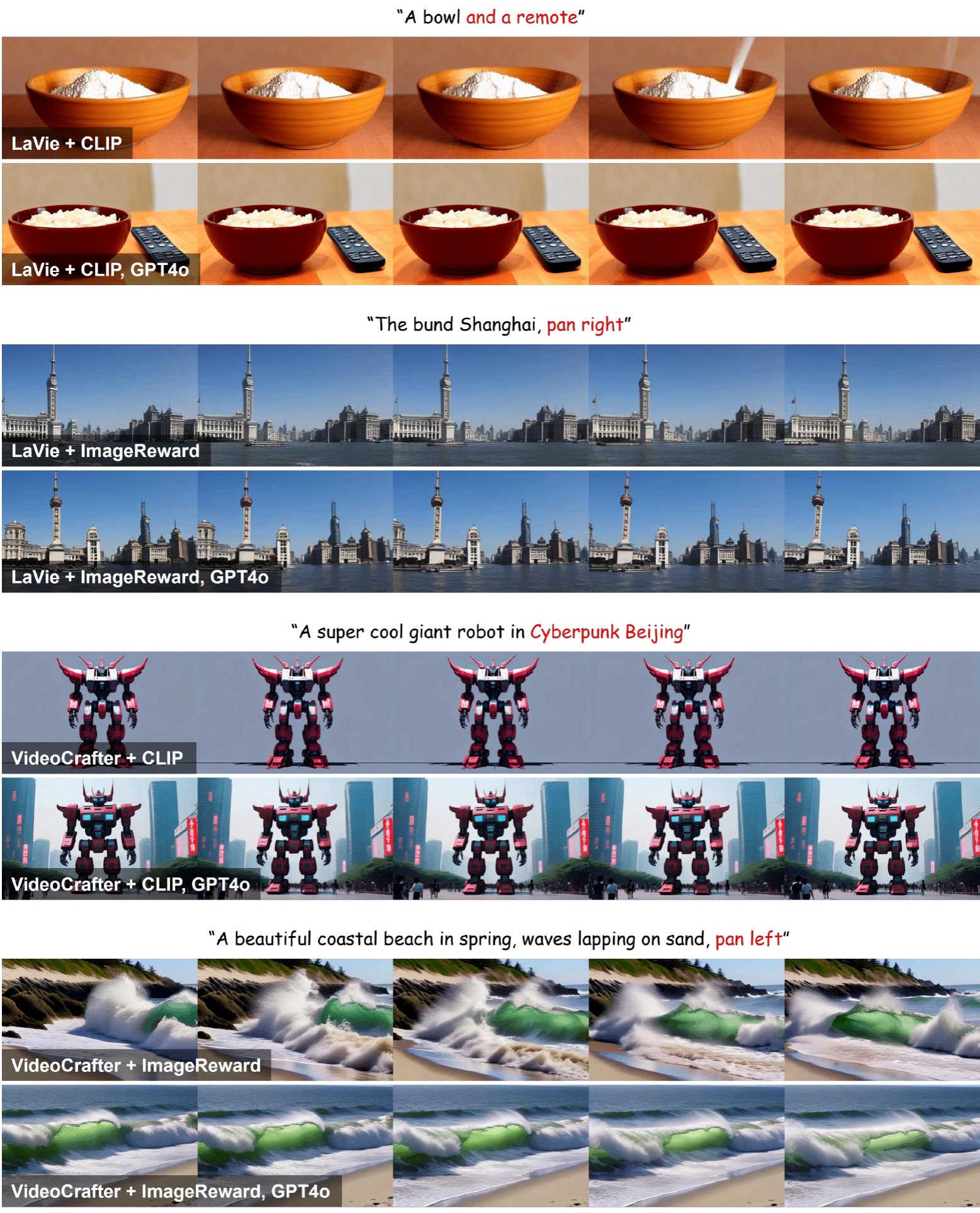}
    \vspace{-0.4cm}
    \caption{More qualitative results of ensembling with LVLMs. The red text highlights the difference between the models.
    }
    \label{fig:results_more2}
    \vspace{-0.2cm}
\end{figure*}

\begin{figure*}[!b]
    \centering
    \includegraphics[width=\linewidth]{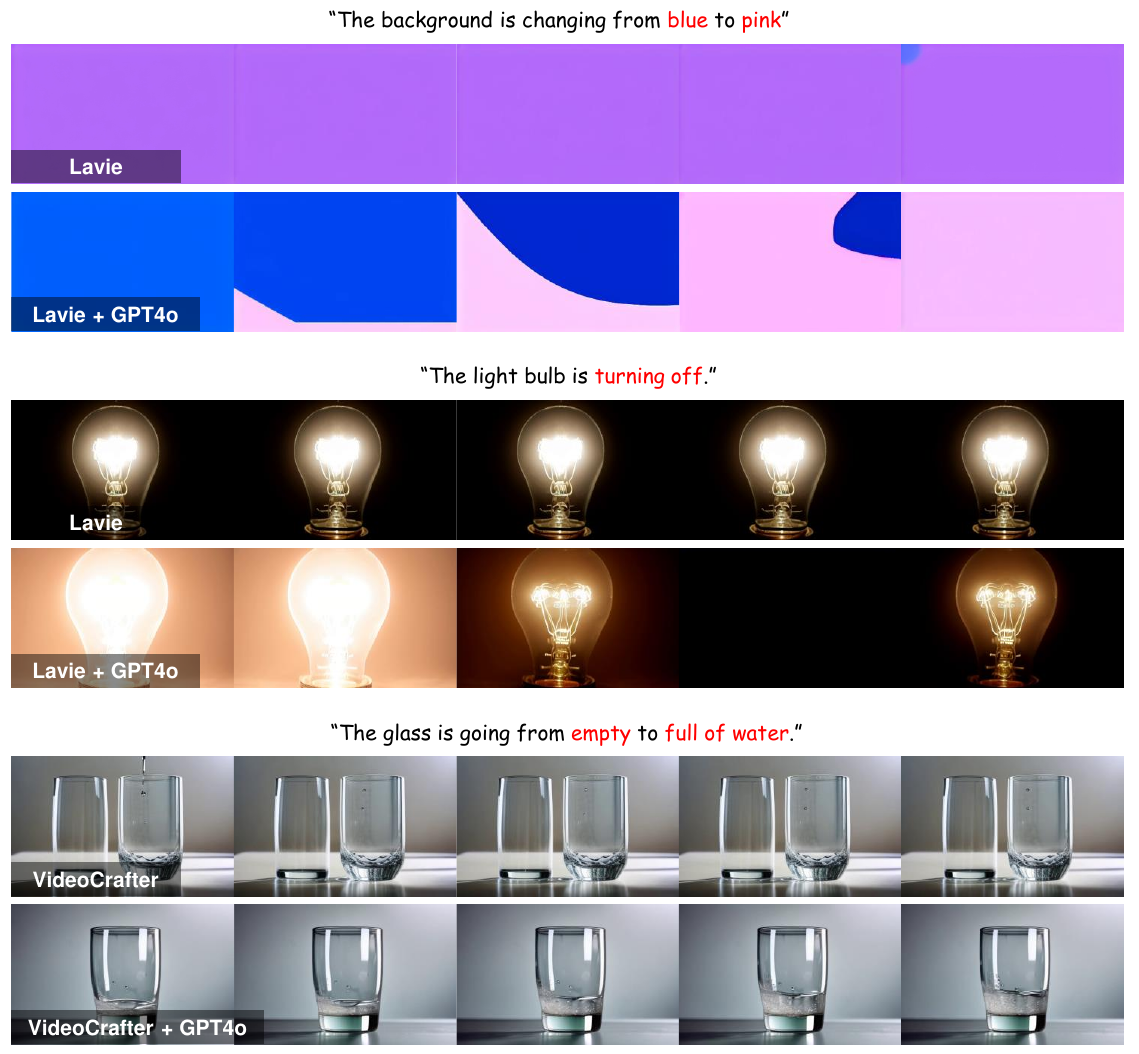}
    \vspace{-0.4cm}
    \caption{More qualitative comparison of T2V-Compbench to analyze video-specific dynamics. The red text highlights the difference between the models.
    }
    \label{fig:results_more1-2}
    \vspace{-0.2cm}
\end{figure*}

\end{document}